\documentclass[10pt,twocolumn,letterpaper]{article}

\usepackage{cvpr}
\usepackage{times}
\usepackage{epsfig}
\usepackage{graphicx}
\usepackage{amsmath}
\usepackage{amssymb}

\usepackage{booktabs}       % professional-quality tables
\usepackage{amsfonts}       %b blackboard math symbols
\usepackage{microtype}      % microtypography
\usepackage{enumitem}
\usepackage{bm}
\usepackage{chngpage}
\usepackage{algorithm}
\usepackage{algorithmic}
\usepackage{subcaption}
\usepackage{wrapfig}
\usepackage{url}            % simple URL typesetting
\usepackage{booktabs}       % professional-quality tables
\usepackage{amsfonts}       % blackboard math symbols
\usepackage{nicefrac}       % compact symbols for 1/2, etc.
\usepackage{microtype}      % microtypography
\usepackage{color, colortbl}
\usepackage[english]{babel}

\DeclareMathOperator*{\argmax}{\arg\!\max}
\usepackage[breaklinks=true,bookmarks=false]{hyperref}

\cvprfinalcopy % *** Uncomment this line for the final submission

\def\cvprPaperID{2905} % *** Enter the CVPR Paper ID here

\ifcvprfinal\fi
\begin{document}

%%%%%%%%% TITLE
\title{Smooth Neighbors on Teacher Graphs for Semi-supervised Learning}

\author{
	Yucen Luo$^{1}$ \hspace{1.2cm}
	Jun Zhu$^{1}$\thanks{Corresponding author.} \hspace{1.2cm}
	Mengxi Li$^{2}$ \hspace{1.2cm}
	Yong Ren$^{1}$ \hspace{1.2cm}
	Bo Zhang$^{1}$\\
	$^{1}$ Dept. of Comp. Sci. \& Tech., State Key Lab for Intell. Tech. \& Sys., BNRist Lab, Tsinghua University\\
	$^{2}$ Department of Electronical Engineering, Tsinghua University\\
	{\tt\small \{luoyc15, limq14, renyong15\}@mails.tsinghua.edu.cn; \{dcszj, dcszb\}@tsinghua.edu.cn}}
\maketitle
\thispagestyle{empty}

%%%%%%%%% ABSTRACT
\begin{abstract}
	The recently proposed self-ensembling methods have achieved promising results in deep semi-supervised learning, which penalize inconsistent predictions of unlabeled data under different perturbations. However, they only consider adding perturbations to each single data point, while ignoring the connections between data samples.
	In this paper, we propose a novel method, called Smooth Neighbors on Teacher Graphs (SNTG). In SNTG,
	%, to explicitly regularize the representations of neighboring data points and enforce more smoothness.
	a graph is constructed based on the predictions of the teacher model, \ie, the implicit self-ensemble of models. Then the graph serves as a similarity measure with respect to which the representations of ``similar'' neighboring points are learned to be smooth on the low-dimensional manifold. We achieve state-of-the-art results on semi-supervised learning benchmarks. The error rates are 9.89\%, 3.99\% for CIFAR-10 with 4000 labels, SVHN with 500 labels, respectively. In particular, the improvements are significant when the labels are fewer. For the non-augmented MNIST with only 20 labels, the error rate is reduced from previous 4.81\% to 1.36\%. Our method also shows robustness to noisy labels.
\end{abstract}
\vspace{-0.2cm}
%%%%%%%%% BODY TEXT
\section{Introduction}
\label{sec:intro}
As collecting a fully labeled dataset is often expensive and time-consuming, semi-supervised learning (SSL) has been extensively studied in computer vision to improve generalization performance of the classifier by leveraging limited labeled data and a large amount of unlabeled data~\cite{chapelle2009semi}.
The success of SSL relies on the key \emph{smoothness} assumption, \ie, data points close to each other are likely to have the same label. It has a special case named \emph{cluster} or \emph{low density separation} assumption, which states that the decision boundary should lie in low density regions, not crossing high density regions~\cite{chapelle2005semi}. Based on these assumptions, many traditional methods have been developed~\cite{joachims1999transductive,zhu2003semi,zhou2004learning,chapelle2005semi,belkin2006manifold}.

Recently due to the great advances of deep learning~\cite{krizhevsky2012imagenet}, remarkable results have been achieved on SSL~\cite{kingma2014semi,rasmus2015semi,salimans2016improved,laine2016temporal}. Among these works, perturbation-based methods~\cite{rifai2011manifold,bachman2014learning,rasmus2015semi,sajjadi2016regularization,laine2016temporal} have demonstrated great promise.
Adding noise to the deep model is important to reduce overfitting and learn more robust abstractions, \eg, dropout~\cite{hinton2012improving} and randomized data augmentation~\cite{ciregan2012multi}. In SSL, perturbation regularization aids by exploring the \emph{smoothness} assumption. For example, the Manifold Tangent Classifier (MTC)~\cite{rifai2011manifold} trains contrastive auto-encoders to learn the data manifold and regularizes the predictions to be insensitive to local perturbations along the low-dimensional manifold. Pseudo-Ensemble~\cite{bachman2014learning} and $\Gamma$ model in Ladder Network~\cite{rasmus2015semi} evaluate the classifiers with and without perturbations, which act as a \emph{``teacher''} and a \emph{``student''}, respectively. The student needs to predict consistently with the targets generated by the teacher on unlabeled data. Following the same principle, temporal ensembling, mean teacher and virtual adversarial training~\cite{laine2016temporal,tarvainen2017mean,miyato2017virtual} improve the target quality in different ways to form better teachers.
%Motivated by those works, our method exploits more information in the teacher and provides additional smoothness in the feature space.
All these approaches aim to fuse the inputs into coherent clusters by adding noise and smoothing the mapping function locally~\cite{laine2016temporal}.

However, these methods only consider the perturbations around \emph{each single} data point, while ignoring the connections between data points, therefore not fully utilizing the information in the unlabeled data structure, such as clusters or manifolds.
An extreme situation may happen where the function is smooth in the vicinity of each unlabeled point but not smooth in the vacancy among them. This artifact could be avoided if the unlabeled data structure is taken into consideration.
It is known that data points similar to each other (\eg, in the same class) tend to form clusters (\emph{cluster} assumption). Therefore, the connections between similar data points help the fusing of clusters become tighter and more effective (see Fig.~\ref{fig:embed} for the visualization of real data).

%\begin{figure}
%	\centering
%	\includegraphics[width=\linewidth]{images/cartoon}\vspace{-.2cm}
%	\caption{An illustration of how SNTG improves the binary classification (triangle and circle). Labeled data are solid and unlabeled data are hollow. \emph{Left}: Without SNTG, features are loosely fused to 3 clusters (marked in colors). The smallest cluster in the middle including three triangles are misclassified to another side of the decision boundary. \emph{Right}: With SNTG, the clusters are fused tighter yield more accurate predictions.}
%	\label{fig:cartoon}\vspace{-.2cm}
%\end{figure}

%We depict the structure of our model in Fig.~\ref{fig:modelstructure1}. Apart from supervised loss on labeled data and consistency loss evaluated on perturbed single points, we construct a teacher graph to induce smoothness for neighboring points in the feature space.
\begin{figure}[t]
	\centering\vspace{-1.48cm}
	\includegraphics[width=1.12\linewidth]{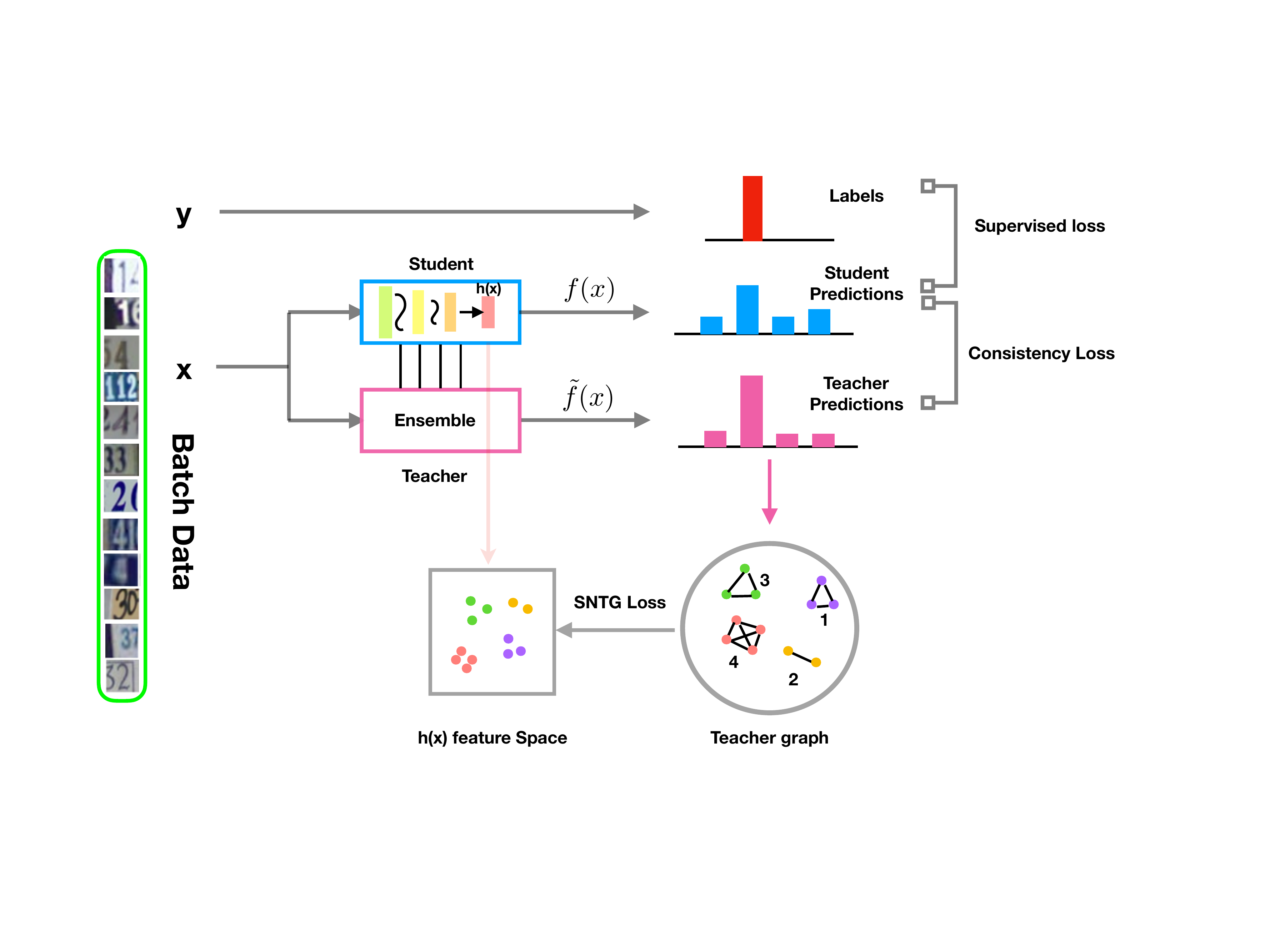}\vspace{-1.6cm}
	\caption{The structure of our model.}\vspace{-.6cm}
	\label{fig:modelstructure}
\end{figure}

Motivated by that, we propose \emph{Smooth Neighbors on Teacher Graphs} (SNTG) that considers the connections between data points to induce smoothness on the data manifold.
% learn neighboring structure to induce smoothness on the data manifold.
By learning a teacher graph based on the targets generated by the teacher,
% to encode the neighborhood relationships,
our model encourages invariance when some perturbations are added to the neighboring points on the graph.
Since deep networks have a hierarchical property, the top layer maps the inputs into a low-dimensional feature space~\cite{bengio2009learning,sharif2014cnn,NIPS2016_6263}. Given the teacher graph, SNTG makes the learned features more discriminative by enforcing them to be similar for neighbors and dissimilar for those non-neighbors. The model structure is depicted in Fig.~\ref{fig:modelstructure}. We then propose a doubly stochastic sampling algorithm to reduce the computational cost with large mini-batch sizes. Our method can be applied with very little engineering effort to existing deep SSL works including both generative and discriminative approaches because SNTG does not introduce any extra network parameters. We demonstrate significant performance improvements over state-of-the-art results while the extra time cost is negligible.% minimal
% By comparing with various strong competitors, we show that SNTG achieves state-of-the-art results on benchmark datasets.

\section{Related work}\vspace{-0.1cm}
\label{sec:related}
Using unlabeled data to improve generalization has a long and rich history and the literature in SSL is vast~\cite{zhu2006semi,chapelle2009semi}.
So in this section we focus on reviewing the closely related papers, especially the recent advances in SSL with deep learning.
%So in this section we focus on the key previous papers\junz{do you mean the not-mentioned papers are not key?} and other recent advances in SSL with deep learning.

Self-training methods iteratively use the current classifier to label those unlabeled ones with high confidence~\cite{rosenberg2005semi}. Co-training~\cite{blum1998combining,nigam2000analyzing} uses a pair of classifiers with disjoint views of data to iteratively learn and generate training labels.
Transductive SVMs~\cite{joachims1999transductive} implement the \emph{cluster} assumption by keeping unlabeled data far away from the decision boundaries. Entropy minimization~\cite{grandvalet2005semi}, a strong regularization term commonly used, minimizes the conditional entropy $H\left(p\left(y|x\right)\right)$ to ensure that one instance is assigned to one class with a high probability to avoid class overlap.

%\textbf{Perturbation Regularization.}
%Adding noise to the deep model is important to reduce overfitting and learn more robust abstractions, \eg, dropout~\cite{hinton2012improving} and randomized data augmentation~\cite{ciregan2012multi}. In SSL, perturbation regularization explores the \emph{smoothness} assumption. The Manifold Tangent Classifier (MTC)~\cite{rifai2011manifold}
%regularizes predictions to be smooth with respect to the manifold underlying the data distribution. MTC trains contrastive auto-encoders to learn the manifold and enforces the outputs to be insensitive to local perturbations along the low-dimensional manifold. Pseudo-Ensemble~\cite{bachman2014learning} and $\Gamma$ model in Ladder Network~\cite{rasmus2015semi} evaluate the classifiers with and without perturbations, which act as ``teacher'' and ``student'', respectively. The student needs to predict consistently with the targets generated by the teacher. Following the same principle, the previously mentioned temporal ensembling, mean teacher and VAT~\cite{laine2016temporal,tarvainen2017mean,miyato2017virtual} improve the target quality in different ways. Motivated by those works, our method exploits more information in the teacher and provides additional smoothness in the feature space.
%%, which makes the model robust against random perturbations.

\textbf{Graph-based Methods.}
Graph-based SSL methods~\cite{zhu2002learning,zhu2003semi,zhou2004learning} define the similarity of data points by a graph and make predictions smooth with respect to the graph structure. Many of them often optimize a supervised loss over labeled data with a graph Laplacian regularizer~\cite{belkin2006manifold,gong2015deformed}. Label propagation~\cite{zhu2002learning} pushes label information from a labeled instance to its neighbors using a predefined distance metric. We emphasize that our work differs from these traditional methods in the construction and utilization of the graph. Previous work usually constructs the graph in advance using prior knowledge or manual labeling and the graph remains fixed in the following training process~\cite{belkin2006manifold, weston2008deep}. This can lead to several disadvantages as detailed in Sec.~\ref{sec:mapping} and~\ref{sec:fixgraph}. Although some works~\cite{wang2013dynamic} establish the graph dynamically during the classification, their performance is far from recent state-of-the-art deep learning based methods.
%For example, \emph{Embed}NN~\cite{weston2008deep} defines the graph based on the distance in input space, which is typically low-level (\eg, pixel values of images). For natural images, pixel distance cannot reflect semantic similarity well. Instead, we use the teacher-generated targets in label space, which is high-level and changing as the model evolves. Detailed comparison results are in Sec.~\ref{sec:mapping} and~\ref{sec:fixgraph}.

\textbf{Generative Approaches.}
Besides aforementioned discriminative approaches, another line is generative models, which pay efforts to learn the input distribution $p(x)$ that is believed to share some information with the conditional distribution $p(y|x)$~\cite{lasserre2006principled}. Traditional models such as Gaussian mixtures~\cite{zhu2006semi} try to maximize the joint log-likelihood of both labeled and unlabeled data using EM.
%, which also implements the \emph{cluster} assumption because a given cluster belongs to the same class.
For modern deep generative models, variational auto-encoder (VAE) makes it scalable by employing variational methods combined with deep learning~\cite{kingma2014semi} while generative adversarial networks (GAN) generate samples by optimizing an adversarial game between the discriminator and the generator~\cite{springenberg2015unsupervised,salimans2016improved,li2017triple,dai2017good}. The samples generated by GAN can be viewed as another kind of ``data augmentation'' to ``tell'' the decision boundary where to lie. For example, ``fake'' samples can be generated in low density regions where the training data is rare~\cite{salimans2016improved,dai2017good} based on the \emph{low density separation} assumption. Alternatively, more ``pseudo'' samples could be generated in high density regions to keep away from the decision boundary thus improve the robustness of the classifier~\cite{li2017triple}. Our work is complementary to these efforts and can be easily combined with them. We observe improvements over feature matching GAN~\cite{salimans2016improved} with SNTG (see Section~\ref{app:gan}).

\section{Background}
\label{sec:problem}
We consider the semi-supervised classification task, where the training set $\mathcal{D}$ consists of $N$ examples, out of which $L$ have labels and the others are unlabeled. Let $\mathcal{L}=\{(x_i, y_i)\}_{i=1}^{L}$ be the labeled set and $\mathcal{U} = \{x_i\}_{i=L+1}^N$ be the unlabeled set where the observation $x_i\in \mathcal{X}$ and the corresponding label $y_i\in \mathcal{Y} = \{1,2,...,K\}$. We aim to learn a function $f:\mathcal{X}\to [0,1]^K$ parameterized by $\theta\in \Theta$ by solving a generic optimization problem:\\[-.3cm] %to minimize the classification loss for the empirical data distribution:
\begin{equation}
\label{equ:2}
\min_{\theta}\sum\limits_{i=1}^L \ell(f(x_i;\theta), y_i) + \lambda R(\theta, \mathcal{L},\mathcal{U}),
\end{equation}\\[-.4cm]
where $\ell$ is a pre-defined loss function like cross-entropy loss and $f(x;\theta)$ represents the predicted distribution $p\left( y|x;\theta\right)$.
Since only a small portion of training data is labeled ($L\ll N$), the regularization term $R$ is important to leverage unlabeled data.
Here, $\lambda$ is a non-negative regularization parameter that controls how strongly the regularization is penalized.

\subsection{Perturbation-based methods}
\label{sec:perturb}
As mentioned earlier, the models in perturbation-based methods assume a dual role, \ie, a teacher and a student~\cite{lopez2015unifying}. The training targets for the student are generated by the teacher. Recent progresses focus on improving the quality of targets by using self-ensembling and exploring different perturbations~\cite{laine2016temporal,tarvainen2017mean,miyato2017virtual}, as summarized in~\cite{tarvainen2017mean}.
Formally, self-ensembling methods~\cite{laine2016temporal,tarvainen2017mean}
%The aforementioned self-ensembling methods~\cite{laine2016temporal,tarvainen2017mean}
fit in Eq.~\eqref{equ:2} by defining $R$ as a consistency loss:\\[-.5cm]
\begin{eqnarray}\vspace{-0.2cm}
\label{equ:3}
%	R_C(\theta, \mathcal{L}, \mathcal{U}) = \sum_{i=1}^{N}\mathbb{E}_{\xi',\xi}\| \tilde f(x_i;\theta', \xi') - f(x_i;\theta, \xi)\|^2 ,
R_C(\theta, \mathcal{L}, \mathcal{U}) = \sum_{i=1}^{N}\mathbb{E}_{\xi',\xi} \;d \big(\tilde f(x_i;\theta', \xi'), f(x_i;\theta, \xi)\big) ,
\end{eqnarray}\\[-.4cm]
where $\tilde{f}$ is a ``noisy'' teacher model with parameters $\theta'$ and random perturbations $\xi'$, similarly, $f$ is a student model with $\theta$ and $\xi$, and $d(\cdot,\cdot)$ denotes the divergence between the two distributions. For example, $d$ can be $l_2$ distance or KL divergence. The perturbations include the input noise and the network dropout. The teacher is defined as an implicit ensemble of previous student models and is expected to give better predictions than the student. $\tilde{f}(x)$ can be seen as the training targets and the student is supposed to predict consistently with $\tilde{f}(x)$. Below are several ways to define the teacher $\tilde{f}$, which have proven effective in previous work~\cite{laine2016temporal,tarvainen2017mean,miyato2017virtual}.

\textbf{$\Pi$ model~\cite{laine2016temporal}.} In order to alleviate the bias in the targets, $\Pi$ model adds noise $\xi'$ to $\tilde f$, which shares the same parameters with $f$, \ie, $\theta' = \theta$ in Eq.~\eqref{equ:3}. $\Pi$ model evaluates the network twice under different realizations of i.i.d. perturbations $\xi'$ and $\xi$ every iteration and minimizes their $l_2$ distance. We observe that, in this case, optimizing the objective in Eq.~\eqref{equ:3} is equivalent to minimizing the variance of the prediction. See details in Appendix~\ref{app:rethink}.

\textbf{Temporal ensembling (TempEns)~\cite{laine2016temporal}.} To reduce the variance of targets, TempEns maintains an exponentially moving average (EMA) of predictions over epochs as $\tilde{f}$. The ensemble output is defined as\\[-.3cm]
\begin{equation}\label{eq:f}
\tilde{F}^{(t)}(x_i) =  \alpha\tilde{F}^{(t-1)}(x_i) + (1-\alpha) f^{(t)}(x_i;\theta,\xi),
\end{equation} %\\[-.45cm]
where $f^{(t)}:\mathcal{X} \to [0,1]^K$ is the prediction given by the current student model at training epoch $t$ and $\alpha$ is the momentum. The target given by $\tilde{f}$ for $x_i$ at epoch $t$ is the debias correction of $\tilde F^{(t)}$, divided by factor $(1-\alpha^t)$, \ie,
$\tilde{f}^{(t)} (x_i) = {\tilde F^{(t)}(x_i)}/{(1-\alpha^t)}$.
Since the target $\tilde{f}(x_i)$ obtained in TempEns is based on EMA, the network only needs to be evaluated once, leading to a speed-up for $\Pi$ model.

\textbf{Mean teacher (MT)~\cite{tarvainen2017mean}.}
Instead of averaging predictions every epoch, MT updates the targets more frequently to form a better teacher, \ie, it averages parameters $\theta$ every iteration:\\[-.5cm]
\begin{equation}
\theta' \leftarrow \alpha \theta' + (1-\alpha)\theta .
\end{equation}
MT provides more accurate targets and enables learning large datasets. It also evaluates the network twice, one by teacher $\tilde{f}(\cdot;\theta',\xi')$ and the other by student $f(\cdot;\theta,\xi)$.

\textbf{Virtual adversarial training (VAT)~\cite{miyato2017virtual}.}
Instead of $l_2$ distance, VAT defines $R$ as the KL divergence between the model prediction and that of the input under adversarial perturbations $\xi'_{adv}$:\\[-.3cm]
\begin{equation}
\label{equ:vat}
R_C(\theta, \mathcal{L},\mathcal{U}) = \sum_{i = 1}^{N} \mathrm{KL}(\tilde f(x_i;\theta)\|f (x_i ;\theta, \xi_{adv}')).
\end{equation}\\[-.4cm]
It is assumed that a model trained under the worst-case (adversarial) perturbations will generalize well~\cite{miyato2017virtual}.
Generally, VAT is also in the framework of self-ensembling in the sense of enforcing consistent predictions. VAT resembles $\Pi$ model but distinguishes itself in the distance metric and the type of perturbations. $\tilde f$ in Eq.~\eqref{equ:vat} can be seen as the teacher model while $f$ with $\xi'_{adv}$ is treated as the student model.

As these methods generate targets themselves, the teacher model is likely to render incorrect targets. However, previous results~\cite{laine2016temporal,tarvainen2017mean} as well as ours (see Sec.~\ref{sec:benchmark} and~\ref{sec:noisy}) suggest that the ``teacher-student'' models converge well and are robust to incorrect labels. They mitigate the hazard by using a better teacher and the balanced trade-off between $\ell$ and $R_C$. The success of these methods can be understood as indirectly exploiting the \emph{low-density separation} assumption because the points near the decision boundaries are prone to alter predictions under perturbations thus have large consistency losses. The explicitly penalized $R_C$ will keep unlabeled data far away from the decision boundaries in low density regions and concentrated in high density regions.

\section{Our approach}
\label{sec:approach}
One common shortcoming of the perturbation-based methods is that they regularize the output to be smooth near a data point locally, while ignoring the cluster structure. We address it by proposing a new SSL method, SNTG, that enforces neighbors to be smooth, which is a stronger regularization than only imposing smoothness at a single unlabeled point.
We show that SNTG contributes to form a better teacher model, which is the focus of recent advances on perturbation-based methods.
In the following, we formalize our approach by answering two key questions: (1) how to define the graph and neighbors? and (2) how to induce the smoothness of neighboring points using the graph?

\subsection{Learning the graph with the teacher model}
\label{sec:graph}
Most existing graph-based SSL methods~\cite{belkin2006manifold,weston2008deep} depend on a distance metric in the input space $\mathcal{X}$, which is typically low-level (\eg, pixel values of images). For natural images, pixel distance cannot reflect semantic similarity well. Instead, we use the distance in the label space $\mathcal{Y}$, and treat the data points from the same class as neighbors. However, an issue is that the true labels of unlabeled data are unknown.
We address it by learning a teacher graph using the targets generated by the teacher model.
Self-ensembling is a good choice for constructing the graph because the ensemble predictions are expected to be more accurate than the outputs of current classifier.
%They can be used as the targets for the student model as shown in the previous section.
Inspired by that, a teacher graph can guide the student model to move in correct directions. A comparison
%between the teacher graph and
to
other graphs could be found in Sec.~\ref{sec:fixgraph}.

Formally, for $x_i \in \mathcal{D}$, a target prediction $\tilde f(x_i)$ is given by the teacher defined in the previous section. Denote the \emph{hard} target as ${\tilde{y}}_{i} = \argmax_k\left[ \tilde{f}(x_i)\right] _k$ where $\left[ \cdot\right] _k$ is the $k$-th component of the vector, indicating the probability that the example is of class $k$. We build the graph as follows:\\[-.3cm]
\begin{equation}\label{eq:W}
W_{ij} = \left \{ \begin{array}{ll}
1 & \text{if} ~ {\tilde{y}}_{i} = {\tilde{y}}_{j}  \\
0 & \text{if} ~ {\tilde{y}}_{i}  \not = {\tilde{y}}_{j}
\end{array} \right . ,
\end{equation}\\[-.3cm]
where $W_{ij}$ measures the similarity between sample $x_i$ and $x_j$ and those pairs with nonzero entries are treated as ``neighbors''. Here we simply restrict $W_{ij} \in \{0,1\}$ to construct a $0$-$1$ sparse graph.
Other choices include computing the KL divergence between the {\it soft} predictions $\tilde f(x_i)$ and $\tilde f(x_j)$.
%$x_i, x_j$ are either labeled or unlabeled data and the ensemble predictions are treated as the target labels.
%For the labeled data, we can use the true labels instead, \ie, replace $\tilde y_i$ with
%$y_i$ for $x_i \in \mathcal{L}$. We find that there is marginal difference empirically since the supervised loss $\ell$ makes the predictions on the labeled data almost the same as their true labels.

\subsection{Guiding the low-dimensional feature mapping}
\label{sec:mapping}
Given a graph, we clarify how to regularize neighbors with smoothness. Generally, a deep classifier (\ie, the student) $f$ can be decomposed as $ f = g \circ h$, where $h:\mathcal{X} \to \mathbb{R}^p$ is the mapping from the input space to the penultimate layer and $g :\mathbb{R}^p \to [0,1 ]^K$ is the output layer usually parameterized by a fully-connected layer with softmax. Due to the hierarchical nature of deep networks, $h(x)$ can be seen as a
low-dimensional feature of the input.
% low-dimensional feature mapped from the input space to a smooth and coherent space.
And the feature space is expected to be linearly separable, as shown in the common practice that a following linear classifier $g$ suffices. In terms of approximating the semantic similarity of two instances, the Euclidean distance of $h(x_i)$ and $h(x_j)$ is more suitable than that of $f(x)$ which represents class probabilities. Hence we use the graph to guide $h(x)$ in the feature space, making them distinguishable among classes.

Given a $N\times N$ similarity matrix $W$ of the sparse graph, we define the SNTG loss as\\[-.2cm]
\begin{equation}
\label{equ:8}
R_{S}(\theta,\mathcal{L},\mathcal{U}) = \sum\limits_{x_i,x_j \in \mathcal{D}} \ell_G(h(x_i;\theta), h(x_j;\theta), W_{ij})
\end{equation}\\[-.2cm]
The choice of $\ell_G$ is quite flexible, which is related to unsupervised feature learning or clustering. Traditional choices include multidimensional scaling~\cite{cox2000multidimensional}, ISOMAP~\cite{tenenbaum2000global} and Laplacian eigenmaps~\cite{belkin2003laplacian}.
Here we utilize the contrastive Siamese networks~\cite{bromley1994signature} since they are able to learn an
invariant mapping to a smooth and coherent feature space and perform well in metric learning and face verification~\cite{hadsell2006dimensionality,chopra2005learning,taigman2014deepface}. Specifically, the loss is defined as follows:\\[-.2cm]
\begin{equation}
\label{marginloss}
\!\ell_G \!= \!\left \{ \begin{array}{ll}
\|h(x_i) - h(x_j) \|^2 							    & \text{if} ~ W_{ij} \!=\! 1\!\\
\max\left(0, m\! -\! \|h(x_i) - h(x_j) \|\right)^2	& \text{if} ~ W_{ij} \!=\! 0\!
\end{array}
\right .
\end{equation}\\[-.2cm]
where $m> 0$ is a pre-defined margin and $\|\cdot\|$ is Euclidean distance. The margin loss is to constrain neighboring points to have consistent features. Consequently, the neighbors are encouraged to have consistent predictions while the non-neighbors (\ie, the points of different classes) are pushed apart from each other with a minimum distance $m$. Visualizations can be found in Section~\ref{sec:vis}.

One interpretation of why the proposed method works well is that SNTG explores more information in the teacher and improves the target quality.
%The student $f$ and the teacher graph facilitate each other.
The teacher graph leads to better abstract representations in a smooth and coherent feature space and then aids the student $f$ to give more accurate predictions. In turn, an improved student contributes to a better teacher model which can provide more accurate targets.
%AsOur method also shows robustness to incorrect labels.
%, giving a new graph encoding the learned neighborhood relationships more precisely.
Another perspective is that SNTG implements the \emph{manifold} assumption for classification which underlies the loss $\ell_G$, \ie, the points of same class are encouraged to concentrate together on sub-manifolds. The perturbation-based methods only keep the decision boundaries far away from each unlabeled data point while our method encourages the unlabeled data points to form tighter clusters, leading the decision boundaries to locate between the clusters.

We discuss the difference between SNTG and two early works LPDGL~\cite{gong2015deformed} and \emph{Embed}NN~\cite{weston2008deep}.
For LPDGL, the definition and the usage of local smoothness are both different from ours. LPDGL defines deformed Laplacian to smooth the predictions of $k$ neighbors in a local region while our work enforces the features to be smooth by the contrastive loss in Eq.~\eqref{marginloss} w.r.t. the 0-1 teacher graph.
%\emph{Embed}NN jointly learns an embedding and the classifier using unlabeled data.
%However, there are several key differences making SNTG unique.
For \emph{Embed}NN, despite they also measure the embedding loss, there are several key differences.
First, inspired by $\Pi$ model, SNTG aims to induce more smoothness using neighbors under perturbations, while \emph{Embed}NN is motivated by using the embedding as an auxiliary task to help supervised tasks and does not consider the robustness to perturbations. Second, \emph{Embed}NN uses a fixed graph $W$ defined by $k$-nearest-neighbor ($k$-NN) based on the distance in $\mathcal{X}$. Our method takes a different approach using the teacher-generated targets in $\mathcal{Y}$. As mentioned in Section~\ref{sec:graph}, the pixel-level distance in $\mathcal{X}$ may not reflect the semantic similarity as well as that in $\mathcal{Y}$ for natural images. Third, once the graph is built in \emph{Embed}NN, the fixed graph cannot leverage the knowledge distilled by the classifier thus cannot be improved any more, while SNTG jointly learns the classifier and the teacher graph as stated above. Furthermore, on the time cost and scalability, SNTG is faster than \emph{Embed}NN and can handle large-scale datasets. $k$-NN in \emph{Embed}NN is slow for large $k$ and even more time-consuming for large-scale datasets. We compute $W$ in the much lower dimensional $\mathcal{Y}$ and use the sub-sampling technique that is to be introduced next. Experimental comparisons are in Section~\ref{sec:fixgraph}.
\begin{algorithm}[t]
	\caption{Mini-batch training of SNTG for SSL}
	\label{alg:1}
	\begin{algorithmic}[1]
		\REQUIRE $x_i=$ training inputs, $y_i$ for labeled inputs in $\mathcal{L}$\\
		\REQUIRE $w(t) = $ unsupervised weight ramp-up function\\
		\REQUIRE $f_\theta(x) = $ neural network with parameters $\theta$
		%		\STATE {\bfseries Input:} $x_i, y_i$\\
		%		$w(t) = $ unsupervised weight ramp-up function\\
		%		$f_\theta(x) = $ neural network with trainable parameters $\theta$
		\FOR {$t$ in [1, numepochs]}
		\FOR {each minibatch $B$}
		\STATE {$f_{i}\leftarrow f_\theta(x_{i\in B})$  evaluate network outputs}
		\STATE {$\tilde{f}_{i}\leftarrow \tilde{f}(x_{i\in B})$ given by the teacher model }
		\FOR { $(x_i, x_j)$ in a minibatch pairs $S$ from $B$}
		\STATE {
			Compute $W_{ij}$ according to Eq.~\eqref{eq:W}
		}
		\ENDFOR
		\STATE {loss $\leftarrow - \frac{1}{|B|} \sum_{i \in (B\cap \mathcal{L})} \log [f_i]_{y_i}$\\
			$+w(t)\Big[\lambda_1 \frac{1}{|B|}\sum_{i\in B} d\big(\tilde f_i,{f}_i\big)$\\
			$\qquad\;+\lambda_2\frac{1}{|S|} \sum_{i,j \in S}\ell_G( h(x_i), h(x_j), W_{ij})\Big]$}
		\STATE {update $\theta$ using optimizers, \eg, Adam~\cite{kingma2014adam}}
		\ENDFOR	
		\ENDFOR
		\STATE{return $\theta$}
	\end{algorithmic}
\end{algorithm}
\label{sec:alg}
\subsection{Doubly stochastic sampling approximation}
Our overall objective is the sum of two components. The first one is the standard cross-entropy loss on the labeled data, and the second is the regularization term, which encourages the smoothness for each single point (\ie, $R_C$) as well as for the neighboring points (\ie, $R_S$). Alg.~\ref{alg:1} presents the pseudo-code. Following~\cite{laine2016temporal}, we use a ramp-up $w(t)$ for both the learning rate and the regularization term in the beginning.

As our model uses deep networks, we train it using Stochastic Gradient Descent (SGD)~\cite{bottou2010large} with mini-batches.
%To reduce variance, SGD-based optimization methods usually adopt mini-batch training.
We follow the common practice and construct the sub-graph in a random mini-batch to estimate $R_S$ in Eq.~\eqref{equ:8}.
For a mini-batch $B$ of size $n$, we need to compute $W_{ij}$ for all the data pairs $(x_i,x_j) \in B$,	which is of size $n^2$ in total. Although this step is fast, the computation of $\|h(x_i) - h(x_j)\|$ related to $W_{ij}$ is $O(p)$ and then the overall computational cost is $O(n^2p)$, which is slow for large $n$. To reduce the computational cost, we instead use doubly stochastic sampled data pairs to construct
$W_{ij}$ and only use them to compute Eq.~\eqref{marginloss}, which is still an unbiased estimation
of $R_S$. Specifically, in each iteration, we sample a mini-batch $B$ and then sub-sample $s \leq n^2$ data pairs $S$ from $B$. Empirically, SNTG can be incorporated into other SSL methods with not much extra time cost.
%Following~\cite{laine2016temporal}, we use a ramp-up $w(t)$ for both learning rate and the regularization term in the beginning and a ramp-down function for annealing the learning rate at the end.
See Appendix~\ref{app:1} for details.

\begin{figure}[t]\vspace{-0.3cm}
	\centering
	\begin{subfigure}{.234\textwidth}
		\centering
		\includegraphics[width=\linewidth]{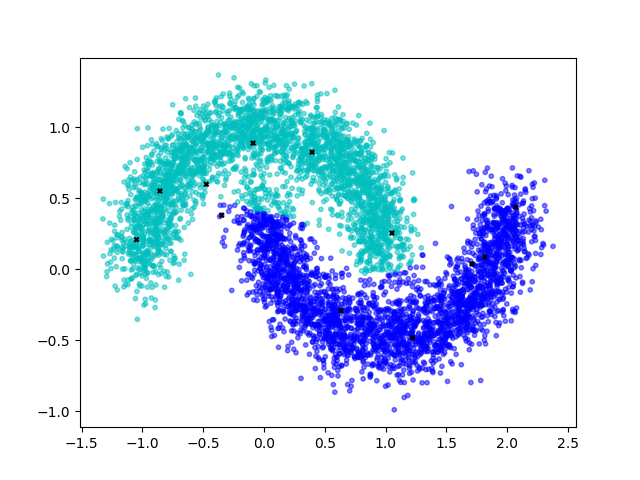}\vspace{-.2cm}
		\caption{``two moons'', $\Pi$ model}
		\label{fig:moon_pi}\vspace{-.1cm}
	\end{subfigure}
	\begin{subfigure}{.234\textwidth}
		\centering
		\includegraphics[width=\linewidth]{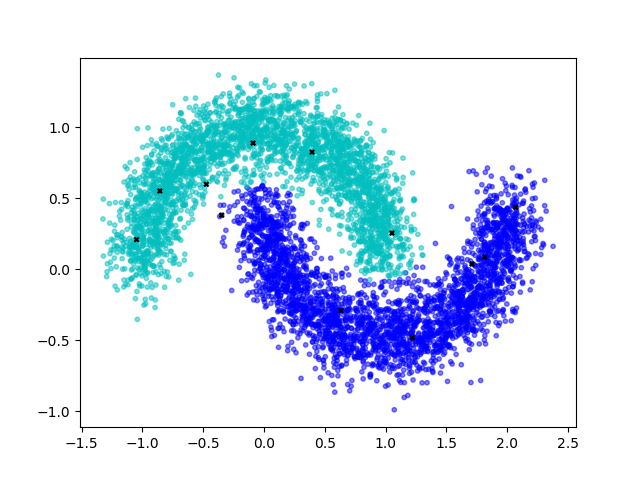}\vspace{-.2cm}
		\caption{``two moons'', SNTG}
		\label{fig:moon_piemb}\vspace{-.1cm}
	\end{subfigure}
	\begin{subfigure}{.234\textwidth}
		\centering
		\includegraphics[width=\linewidth]{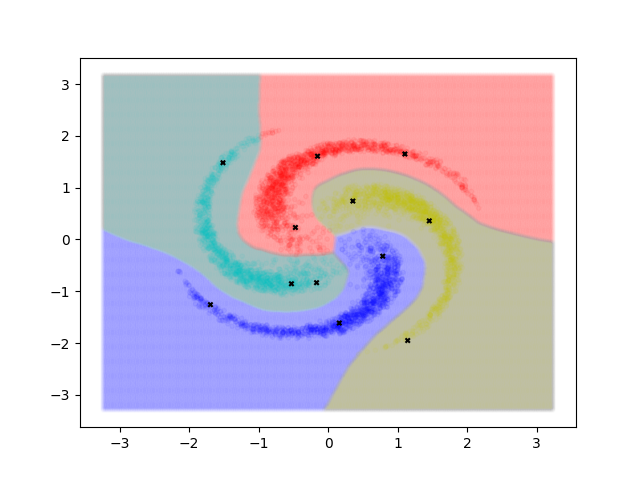}\vspace{-.2cm}
		\caption{``four spins'', $\Pi$ model}
		\label{fig:spin_pi}
	\end{subfigure}
	\begin{subfigure}{.234\textwidth}
		\centering
		\includegraphics[width=\linewidth]{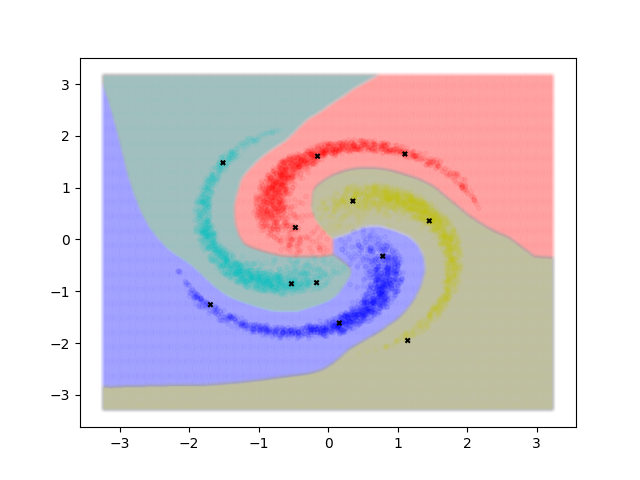}\vspace{-.2cm}
		\caption{``four spins'', SNTG}
		\label{fig:spin_piemb}
	\end{subfigure}\vspace{-.2cm}
	\caption{Comparison between $\Pi$ model (a,c) and SNTG (b,d) on two synthetic datasets. The labeled data are marked with the black cross. Different colors denote different classes. The decision boundaries are shown for \ref{fig:spin_pi} and \ref{fig:spin_piemb}.}
	\label{fig:synthetic}\vspace{-.2cm}
\end{figure}

\section{Experiments}
\label{sec:exp}
This section presents both quantitative and qualitative results to demonstrate the effectiveness of SNTG. The purpose of experiments is to show the improvements that come from SNTG, using cutting-edge approaches as evidence. \footnote{Source code is at \url{https://github.com/xinmei9322/SNTG}.}
%Code will be made available to reproduce all the results after the double-blind review.

%\vspace{-0.2cm}
\subsection{Synthetic datasets}
We first test on the well-known ``two moons'' and ``four spins'' synthetic datasets where $x\in \mathbb{R}^2$ and $y\in\{1,2\}$ and $y\in\{1,2,3,4\}$, respectively. Each dataset includes 6000 data points and the label ratio is 0.002 (\ie, only 12 data points are labeled).
We use neural networks with three hidden layers, each of size 100 with leaky ReLU $\alpha=0.1$ as suggested in CatGAN~\cite{springenberg2015unsupervised}. See Appendix~\ref{app:1} for details.
The results are visualized in Fig.~\ref{fig:synthetic}, where we compare with $\Pi$ model, a strong baseline that performs well with only some failures. Specifically, in Fig.~\ref{fig:moon_pi}, a small blob of data is misclassified to green and in Fig.~\ref{fig:spin_pi}, the tail of the green spin is misclassified to red. The prediction of $\Pi$ model is supposed to be smooth enough at these areas because the data points are in blobs. However, the $\Pi$ model still fails to identify them. For our SNTG, the classifications are both correct in Fig.~\ref{fig:moon_piemb} and Fig.~\ref{fig:spin_piemb} due to effective utilization of neighboring points' structure. Compared to Fig.~\ref{fig:spin_pi}, the decision boundaries in Fig.~\ref{fig:spin_piemb} also align better with the spins. These experiments demonstrate the effectiveness of SNTG.

\begin{table}[t]
	\caption{Error rates (\%) on benchmark datasets without augmentation, averaged over $10$ runs.}
	\label{noaug}\vspace{-.2cm}
	\centering
	\resizebox{.48\textwidth}{!}{ 	
		\begin{tabular}{p{0.3\linewidth}p{0.19\linewidth}p{0.17\linewidth}p{0.18\linewidth}p{0.199\linewidth}}
			\toprule
			Models			&MNIST ($L$=100)				 & SVHN ($L$=1000)			& CIFAR-10 ($L$=4000) & CIFAR-100 ($L$=10000)\\
			\midrule
			%		Supervised-only & 						 &20.47$\pm$2.64		&35.56$\pm$1.59\\
			LadderNetwork~\cite{rasmus2015semi}	&0.89$\pm$0.50			 &--						&20.40$\pm$0.47&--\\
			CatGAN~\cite{springenberg2015unsupervised}			&1.39$\pm$0.28			 &--						&19.58$\pm$0.58&--\\
			ImprovedGAN~\cite{salimans2016improved} 	&0.93$\pm$0.065 		 &8.11$\pm$1.3			&18.63$\pm$2.32&--\\
			ALI~\cite{dumoulin2016adversarially}				&--						 &7.42$\pm$0.65 		&17.99$\pm$1.62&--\\
			TripleGAN~\cite{li2017triple}		&0.91$\pm$0.58			 & 5.77$\pm$0.17 		&16.99$\pm$0.36&--\\
			GoodBadGAN~\cite{dai2017good} 		&0.795$\pm$0.098		 &4.25$\pm$0.03&14.41$\pm$0.03&--\\
			
			\midrule
			$\Pi$ model~\cite{laine2016temporal}		&0.89$\pm$0.15*			 &5.43$\pm$0.25			&16.55$\pm$0.29&39.15$\pm$0.36\\
			$\Pi$+SNTG (\textbf{ours})		&\textbf{0.66$\pm$0.07}	 &4.22$\pm$0.16			&13.62$\pm$0.17&\textbf{37.97$\pm$0.29}\\
			\midrule
			VAT~\cite{miyato2017virtual} 			&	1.36				 &    5.77  			&  	14.82&--\\
			VAT+Ent~\cite{miyato2017virtual}			&	--					 &    4.28  		 	&  13.15&--\\
			VAT+Ent+SNTG (\textbf{ours}) 	&--	&\textbf{4.02$\pm$0.20}&\textbf{12.49$\pm$0.36}&--\\
			\bottomrule
		\end{tabular}\vspace{-.5cm}
	}
\end{table}
\begin{table*}[t]\vspace{-.5cm}
	\caption{Error rates (\%) on SVHN with translation augmentation, averaged over $10$ runs.}\vspace{-.1cm}
	\label{svhn}
	\centering
	\resizebox{.65\textwidth}{!}{ 	
	\begin{tabular}{lllll}
		\toprule
		Model     & 250 labels   & 500 labels    & 1000 labels     & All labels \\
		\midrule
		Supervised-only~\cite{tarvainen2017mean} &42.65$\pm$2.68&22.08$\pm$0.73 &14.46$\pm$0.71&2.81$\pm$0.07\\
		%		Improved GAN~\cite{salimans2016improved}    &	-&18.44$\pm$4.8&8.11$\pm$1.3&	-\\	
		\midrule
		$\Pi$ model~\cite{laine2016temporal}&9.93$\pm$1.15*& 6.65$\pm$0.53 &4.82$\pm$0.17&2.54$\pm$0.04\\
		$\Pi$+SNTG (\textbf{ours})&5.07$\pm$0.25&4.52$\pm$0.30&\textbf{3.82$\pm$0.25}&\textbf{2.42$\pm$0.05}\\
		\midrule
		TempEns~\cite{laine2016temporal}    &12.62$\pm$2.91*		&  5.12$\pm$0.13  	&4.42$\pm$0.16 	& 2.74$\pm$0.06\\
		TempEns+SNTG (\textbf{ours}) &5.36$\pm$0.57 & {4.46$\pm$0.26}&  3.98$\pm$0.21& 2.44$\pm$0.03  \\
		\midrule
		MT~\cite{tarvainen2017mean}   		&4.35$\pm$0.50	& 4.18$\pm$0.27  	& 3.95$\pm$0.19	& 2.50$\pm$0.05\\
		MT+SNTG (\textbf{ours}) & \textbf{4.29$\pm$0.23}& \textbf{3.99$\pm$0.24} & 3.86$\pm$0.27  & 2.42$\pm$0.06  \\
		\midrule
		VAT~\cite{miyato2017virtual}   		&      	-- 		& -- & 5.42 & -- \\
		VAT+Ent~\cite{miyato2017virtual}   		&   --   	 	& -- & 3.86 & --\\
		VAT+Ent+SNTG (\textbf{ours})   		&   --   	 	& -- & 3.83$\pm$0.22 & -- \\
		\bottomrule
	\end{tabular}
}
	\vspace{-.1cm}
\end{table*}
\begin{table*}[t]
	\caption{Error rates (\%) on CIFAR-10 with standard augmentation, averaged over $10$ runs.}
	\label{cifar}\vspace{-.2cm}
	\centering
	 \resizebox{.65\textwidth}{!}{ 	
	\begin{tabular}{lllll}
		\toprule
		Model     								&1000 labels    &2000 labels    &4000 labels     &All labels \\
		\midrule
		Supervised-only~\cite{laine2016temporal}					&--			&--				&34.85$\pm$1.65  &6.05$\pm$0.15\\
		%		Improved GAN 					    &21.83$\pm$2.01 &19.61$\pm$2.09 &18.63$\pm$2.32&--\\
		\midrule
		$\Pi$ model~\cite{laine2016temporal}&31.65$\pm$1.20*&17.57$\pm$0.44*&12.36$\pm$0.31  &5.56$\pm$0.10\\
		$\Pi$+SNTG (\textbf{ours})			&21.23$\pm$1.27	&14.65$\pm$0.31	&11.00$\pm$0.13 &\textbf{5.19$\pm$0.14}\\
		\midrule
		TempEns~\cite{laine2016temporal}	&23.31$\pm$1.01*&15.64$\pm$0.39*&12.16$\pm$0.24  &5.60$\pm$0.10\\
		TempEns+SNTG (\textbf{ours})		&\textbf{18.41$\pm$0.52}&\textbf{13.64$\pm$0.32} &10.93$\pm$0.14 &{5.20$\pm$0.14}\\
		\midrule
		VAT~\cite{miyato2017virtual}    	&--   	 		& -- 			& 11.36 		 &5.81 \\
		VAT+Ent~\cite{miyato2017virtual}   	&--	 			& -- 			& 10.55 		 & -- \\
		%		\rowcolor{Gray}
		VAT+Ent+SNTG (\textbf{ours})					&--   	 		& --			& \textbf{9.89$\pm$0.34}& --\\
		\bottomrule
	\end{tabular}
	}
	\vspace{-.3cm}
\end{table*}
\subsection{Benchmark datasets}
\label{sec:benchmark}

We then provide results on the widely adopted benchmarks, MNIST, SVHN, CIFAR-10 and CIFAR-100. Following common practice~\cite{rasmus2015semi,salimans2016improved}, we randomly sample 100, 1000 4000 and 10000 labels for MNIST, SVHN, CIFAR-10 and CIFAR-100, respectively. We further explore fewer labels for the non-augmented MNIST as well as SVHN and CIFAR-10 with standard augmentation. The results are averaged over 10 runs with different seeds for data splits. Main results are presented in Tables~\ref{noaug},~\ref{svhn},~\ref{cifar} and~\ref{mnist}. The accuracy of baselines are all taken from existing literature. In general, we can see that our method surpasses previous state-of-the-arts by a large margin.

All models are trained with the same network architecture and hyper-parameters to our baselines, \ie, perturbation-based methods described in Sec.~\ref{sec:perturb}. The SNTG loss only needs three extra hyper-parameters: the regularization parameter $\lambda_2$, the margin $m$ and the number of sub-sampled pairs $s$. We fix $m$ and $s$, and only tune $\lambda_2$. More details on experimental setup can be found in Appendix~\ref{app:1}. For fair comparison, we also report our best implementation under the settings not covered in ~\cite{laine2016temporal} (marked $*$).

Note that VAT is a much stronger baseline than $\Pi$ model and TempEns since it explores adversarial perturbation with extra efforts and more time. VAT's best results are achieved with an additional entropy minimization (Ent) regularization ~\cite{grandvalet2005semi}. We evaluate our method under the best setting VAT+Ent and observe a further improvement with SNTG, \eg, from 13.15\% to 12.49\% and from 10.55\% to 9.89\% on CIFAR-10 without or with augmentation, respectively.
In fact, we observed that Ent could also improve the performance of other self-ensembling methods if it was added along with SNTG. But to keep the results clear and focus on the efficacy of SNTG, we did not illustrate the results here.
%Moreover, VAT only reports mean error but omits standard deviation (std) bar, yet another important indicator of the robustness of algorithms. We additionally report the std values along with mean errors over 10 runs.

As shown in Tables~\ref{svhn} and~\ref{cifar}, when SNTG is applied to the fully supervised setting (\ie, all labels are observed), our method further reduces the error rates compared to self-ensembling methods, \eg, from $5.56\%$ to $5.19\%$ on CIFAR-10 for $\Pi$ model. It suggests that supervised learning also benefits from the additional smoothness and the learned invariant feature space in SNTG.

\begin{table}
	\caption{Error rates (\%) on MNIST without augmentation.}% over $10$ runs.}
	\label{mnist}\vspace{-.2cm}
	\centering
	\resizebox{.48\textwidth}{!}{ 	
		\begin{tabular}{p{0.30\linewidth}p{0.19\linewidth}p{0.19\linewidth}p{0.19\linewidth}}
			\toprule
			Models				&20 labels				 	&50 labels			    &100 labels\\
			\midrule
			ImprovedGAN~\cite{salimans2016improved} 		&16.77$\pm$4.52		 		&2.21$\pm$1.36			&0.93$\pm$0.065\\
			Triple GAN~\cite{li2017triple}			&4.81$\pm$4.95			 	& 1.56$\pm$0.72			&0.91$\pm$0.58\\
			$\Pi$ model~\cite{laine2016temporal}			&6.32$\pm$6.90*			 	&1.02$\pm$0.37*			&0.89$\pm$0.15* \\
			\midrule
			$\Pi$+SNTG (\textbf{Ours})	&\textbf{1.36$\pm$0.78}		&\textbf{0.94$\pm$0.42}	&{\textbf{0.66$\pm$0.07}}\\
			\bottomrule
		\end{tabular}
	}
	\vspace{-.35cm}
\end{table}

\textbf{Fewer labels.} Notably, as shown in Tables~\ref{mnist}, \ref{svhn} and \ref{cifar}, when labels are very scarce, \eg, MNIST with 20 labels (only 2 labeled samples per class), SVHN with 250 labels and CIFAR-10 with 1000 labels, the benefits provided by SNTG are even more significant. The SNTG regularizer empirically reduces the overfitting on the small set of labeled data and thus yields better generalization.
%Since the labeled data only accounts for a small part, adding a strong regularizer SNTG in this case helps the model learn faster by exploring the unlabeled data structure.

\begin{table}[t]\vspace{-.05cm}
	\caption{Ablation study on CIFAR-10 with 4000 labels without augmentation. $L_S$ denotes the supervised loss (the first term in Eq.~\ref{equ:2}), and $R_C$ and $R_S$ are defined in text. $L_S$+$R_C$ equals to $\Pi$ model and $L_S$+$R_C$+$R_S$ equals to $\Pi$+SNTG.}\vspace{-.2cm}
	\label{ablation}
	\centering
	\begin{tabular}{lllll}
		\toprule
		Settings 	&$L_S$	&$L_S$+$R_C$	&$L_S$+$R_S$	&$L_S$+$R_C$+$R_S$\\
		\midrule
		Error (\%) 	&35.56	&16.55		&15.36		&13.62\\
		\bottomrule
	\end{tabular}\vspace{-.4cm}
\end{table}
\textbf{Ablation study.}
Our reported results are based on adding SNTG loss $R_S$ to baselines, and the overall objective has already included the consistency loss $R_C$ (See Alg.~\ref{alg:1}, line 9). To quantify the effectiveness of our method, Table~\ref{ablation} presents the evaluation of $\Pi$+SNTG compared to its ablated versions. The error rate of $\Pi$ model, which only uses $R_C$, is $16.55\%$. However, using $R_S$ alone yields a lower error rate of $15.36\%$. Thus, $R_S$ considering the neighbors proves to be a strong regularization, comparable or even favorable to $R_C$, and they are also complementary.

\textbf{Convergence.}
A potential concern of our method is the convergence, since the information in a teacher graph is
likely to be inaccurate at the beginning of training. However, we did not observe any divergent cases in all experiments.
Empirically, the teacher model is usually a little better than the student in training.
%Therefore the teacher graph guides the student in a correct direction.
Furthermore, the ramp-up $w(t)$ is used to balance the trade-off between the supervised loss and regularization, which is important for the convergence as described in previous works~\cite{laine2016temporal,tarvainen2017mean}.
Using the ramp-up weighting mechanism, the supervised loss dominates the learning in earlier training.
As the training continues, the student model has more confidence in the information given by the teacher model, \ie, the target predictions and the graph, which gradually contributes more to the learning process. Fig.~\ref{fig:fixed} shows that our model converges well.

\subsection{Comparison to \emph{Embed}NN and other graphs}
\label{sec:fixgraph}

\begin{figure}[h]\vspace{-.4cm}
	\centering
	\includegraphics[width=0.83\linewidth]{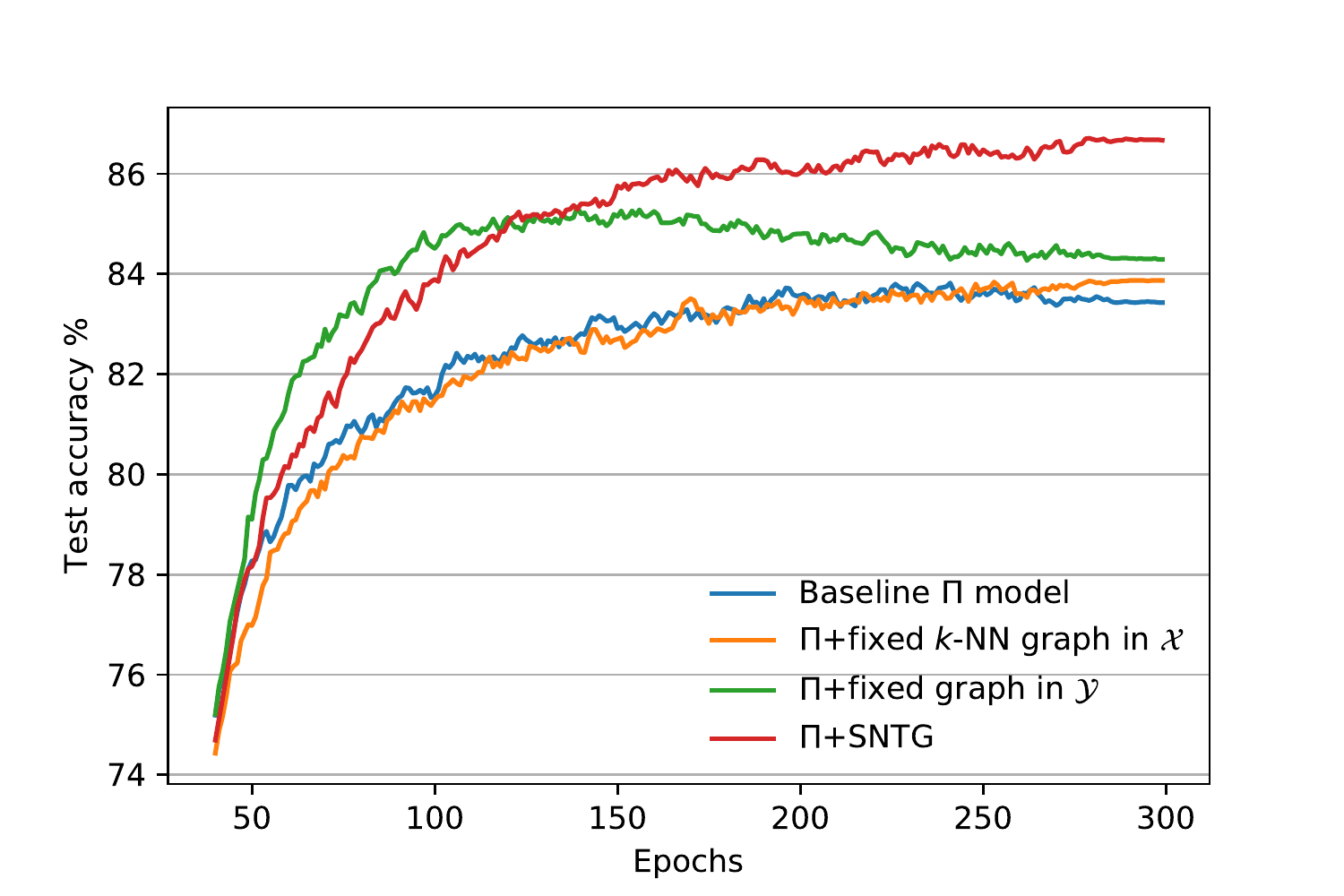}\vspace{-.3cm}
	\caption{Comparison to the fixed graphs on CIFAR-10 with 4000 labels without augmentation.
	}\vspace{-.4cm}
	\label{fig:fixed}
\end{figure}

As our graph is learned based on the predictions in $\mathcal{Y}$ given by the teacher model, we further compare to other graphs. We test them on CIFAR-10 using 4000 labels without augmentation and share all the same hyper-parameter settings with $\Pi$ model except the definition of $W$.
The first baseline is a fixed graph defined by $k$-NN in $\mathcal{X}$---Following \emph{Embed}NN~\cite{weston2008deep}, $W$ is predefined so that 10 nearest neighbors of $x_i$ have $W_{ij}=1$, and $W_{ij}=0$ otherwise.
The second one is another fixed graph in $\mathcal{Y}$---Since only a small portion of labels are observed on training data in SSL, we construct the graph based on the predictions of a pre-trained $\Pi$ model on training data.
Fig.~\ref{fig:fixed} shows that our model outperforms other graphs. The test error rate of the baseline $\Pi$ model is $16.55\%$. Using $k$-NN in $\mathcal{X}$ gives a marginal improvement to $16.13\%$.
Using the predictions in pre-trained $\Pi$ model to construct a 0-1 fixed graph, the error rate is $15.71\%$. Using our method, learning a teacher graph from scratch, $\Pi$+SNTG achieves superior result with $13.62\%$ error rate.

Note that $\Pi$ model is a strong baseline surpassing most previous methods. For natural images like CIFAR-10, the pixel-level distance provides limited information for the similarity thus $k$-NN graph in $\mathcal{X}$ does not improve the strong baseline. The reason of the performance gap to the second one lies in that using a fixed graph in $\mathcal{Y}$ is more like ``pre-training'' while using teacher graph is like ``joint-training''. The teacher graph becomes better using the information extracted by the teacher and then benefits it in turn. However, the fixed graphs cannot receive feedbacks from the model in the training and all the information is from the pre-training or prior knowledge. Empirical results support our analysis.

%t-SNE perplexity 36, learning rate 10, iteration all 702

\begin{figure*}[t]\vspace{-.5cm}
	\centering
	\begin{subfigure}{.33\textwidth}
		\includegraphics[width=\linewidth]{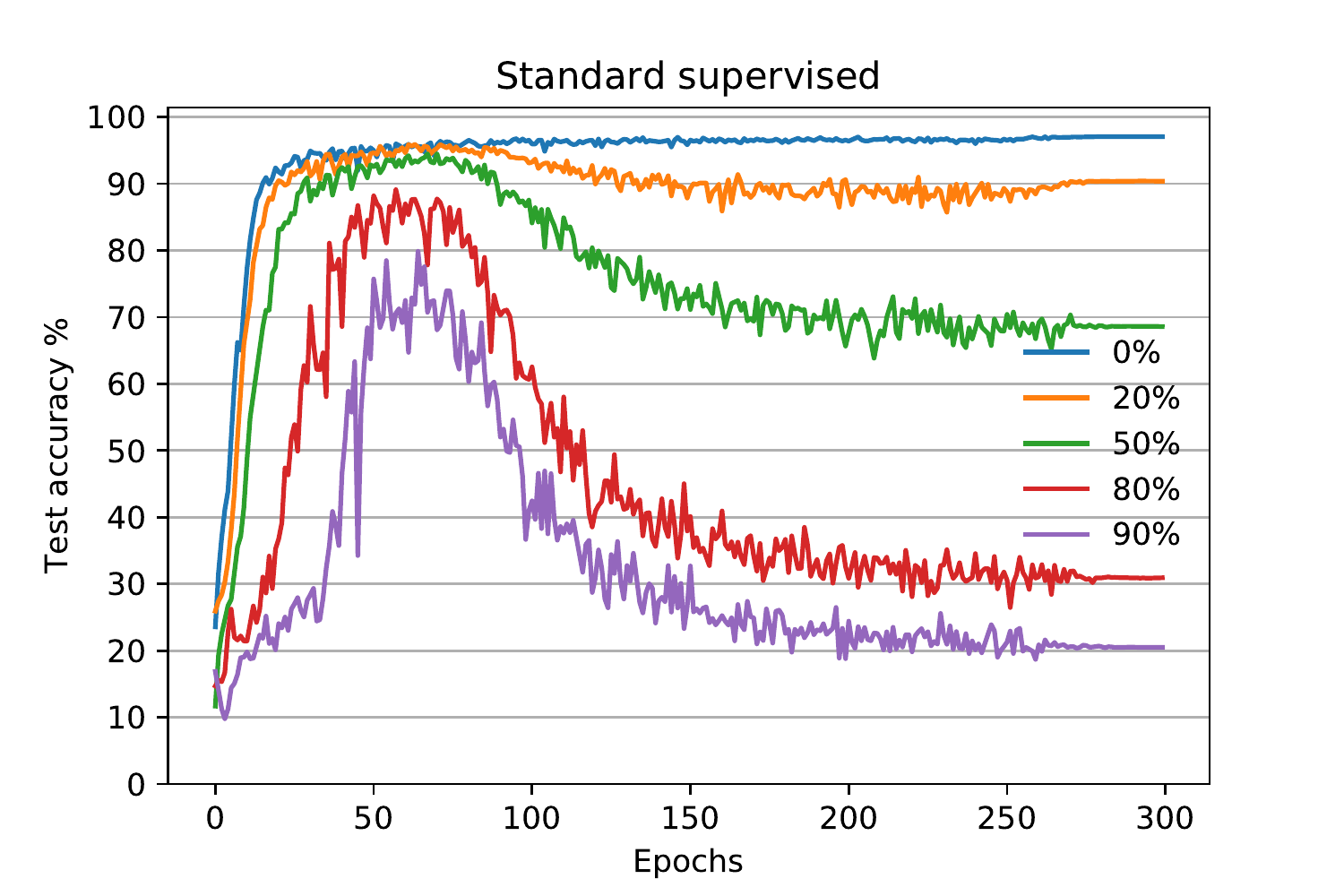}
		\label{fig:corrupt_su}
	\end{subfigure}
	\begin{subfigure}{.33\textwidth}
		\includegraphics[width=\linewidth]{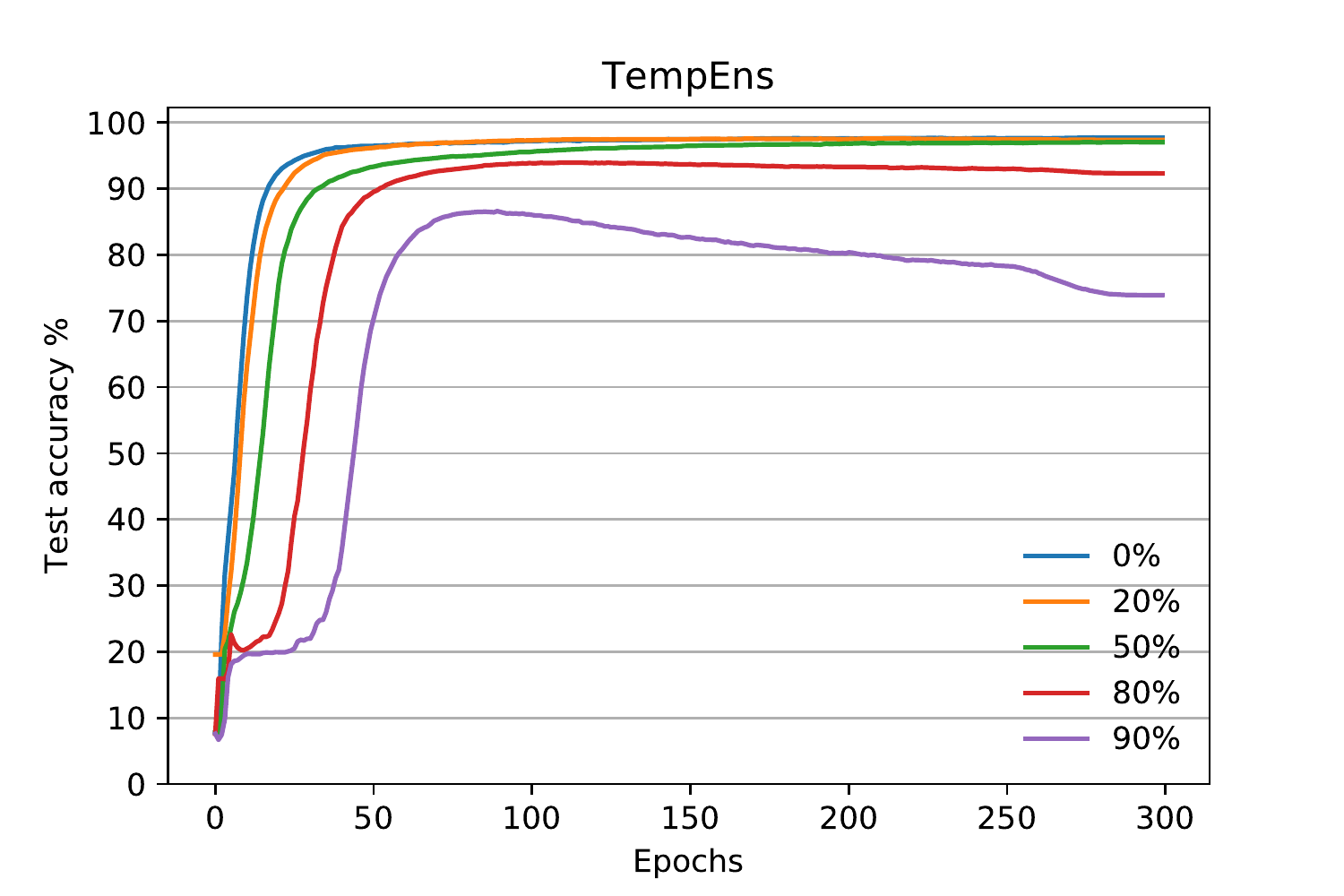}
		\label{fig:corrupt_te}
	\end{subfigure}
	\begin{subfigure}{.33\textwidth}
		\centering
		\includegraphics[width=\linewidth]{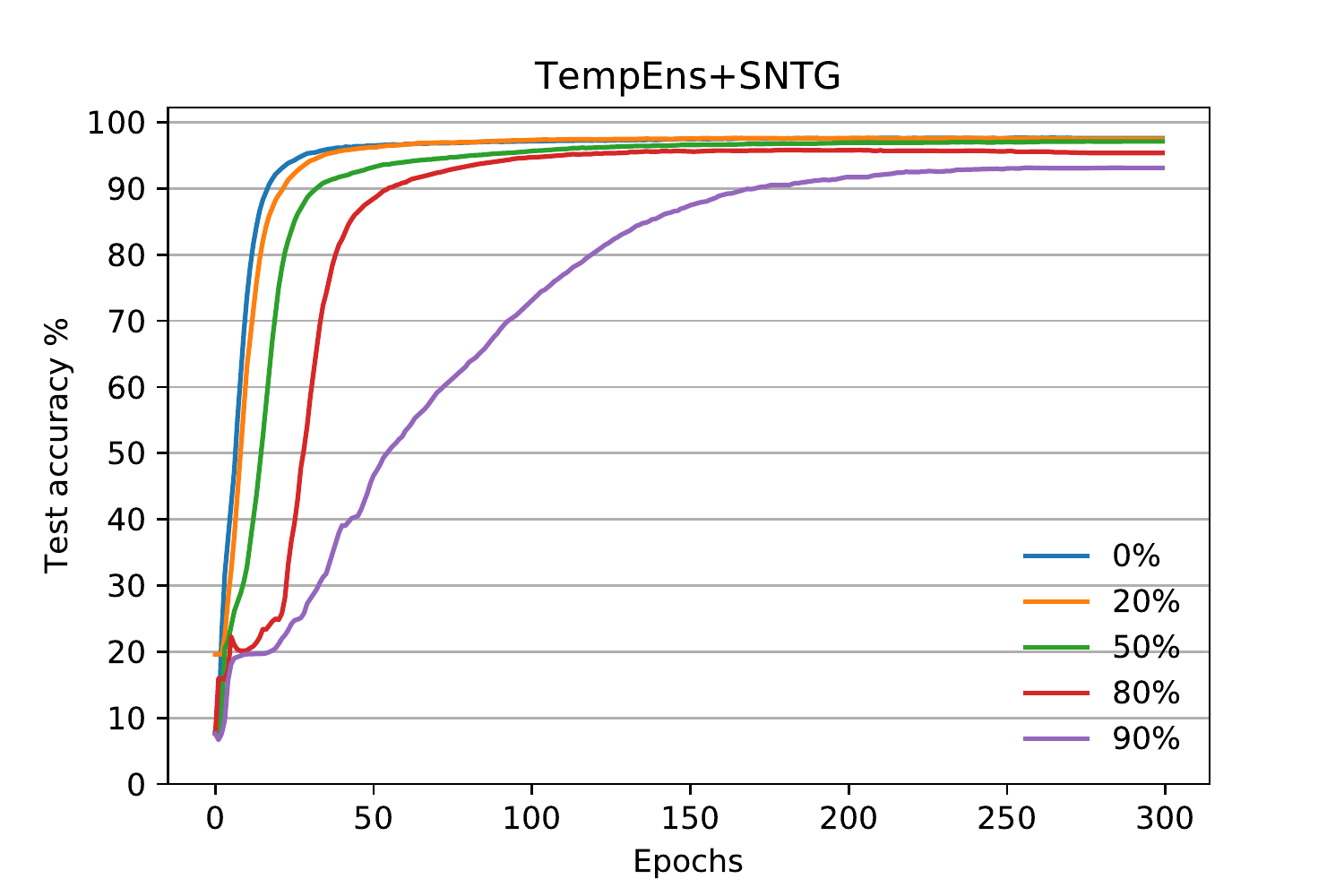}
		\label{fig:corrupt}
	\end{subfigure}\vspace{-.6cm}
	\caption{Test accuracy on supervised SVHN with noisy labels.
		%for three models: (left) standard supervised training; (middle) TempEns; and (right) TempEns+SNTG.
		Different colors denote the percentages of corrupted labels. With standard supervised training (left), the model suffers a lot and overfits to the incorrect information in labels. TempEns (middle) shows the resistance to the corruption but still has a drop in accuracy when the portion of randomized labels increases to 90\%. Adding SNTG shows almost perfect robustness even when 90\% labels are corrupted.
	}\vspace{-.4cm}
	\label{fig:corrupt_all}
\end{figure*}

\vspace{-.1cm}
\subsection{Visualization of embeddings}
\label{sec:vis}
\vspace{-.1cm}
\begin{figure}[t]\vspace{-.3cm}
	\centering
	\begin{subfigure}{.234\textwidth}
		\centering
		\includegraphics[width=\linewidth]{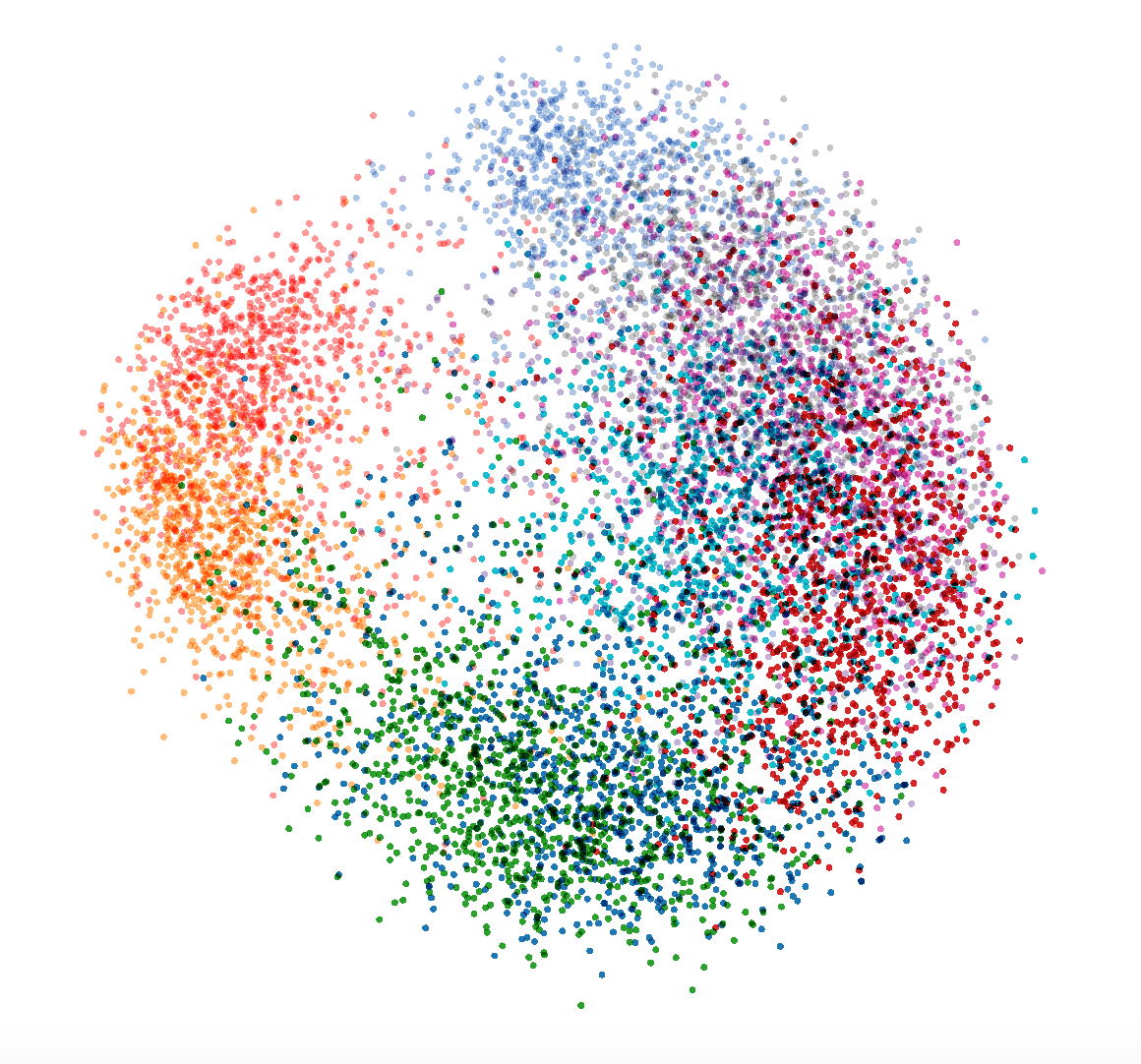}\vspace{-.2cm}
		\caption{CIFAR-10, $\Pi$ model}
		\label{fig:sub1}
	\end{subfigure}
	\begin{subfigure}{.234\textwidth}
		\centering
		\includegraphics[width=\linewidth]{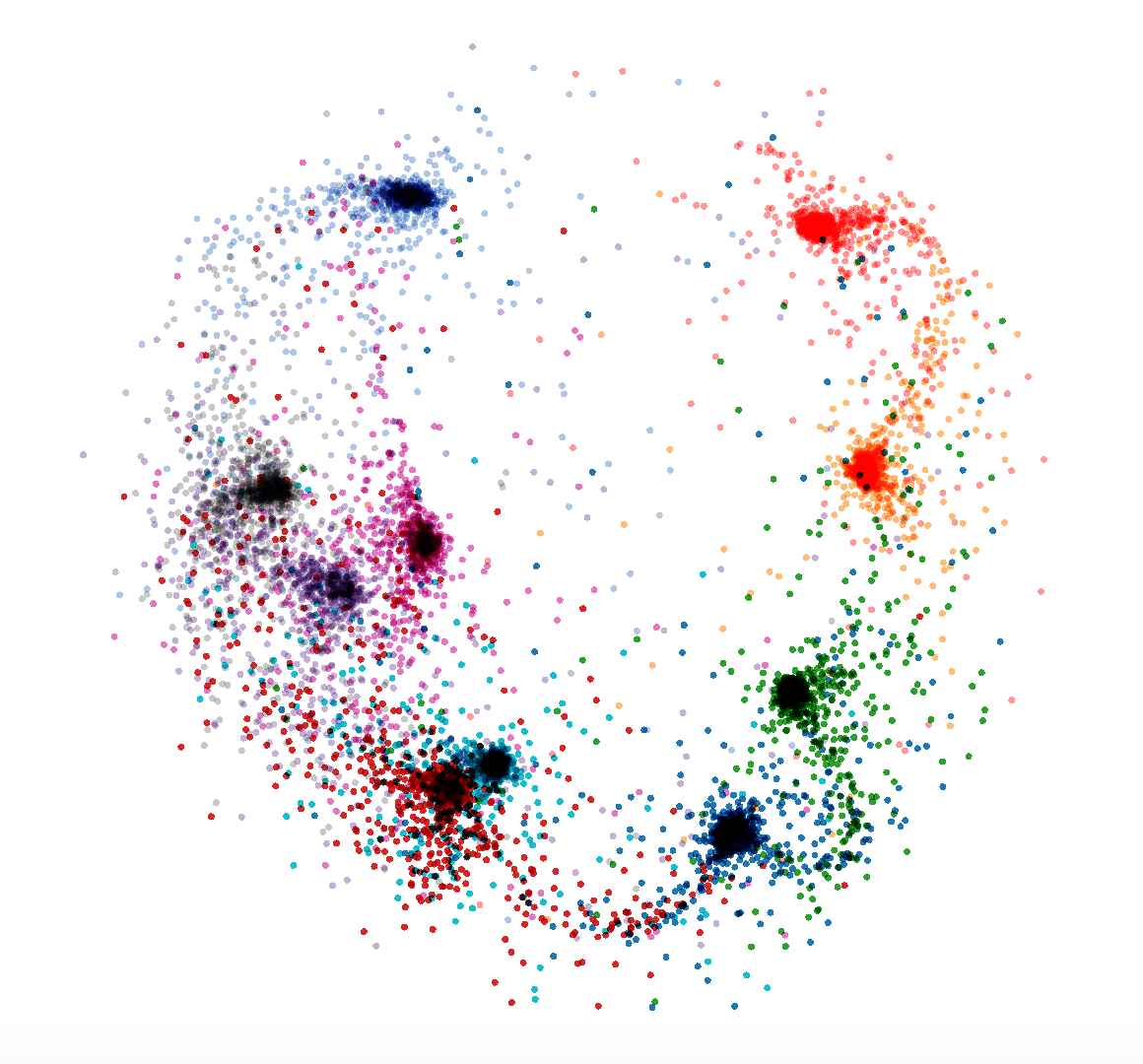}\vspace{-.2cm}
		\caption{CIFAR-10, SNTG}
		\label{fig:sub2}
	\end{subfigure}
	\begin{subfigure}{.234\textwidth}
		\centering
		\includegraphics[width=\linewidth]{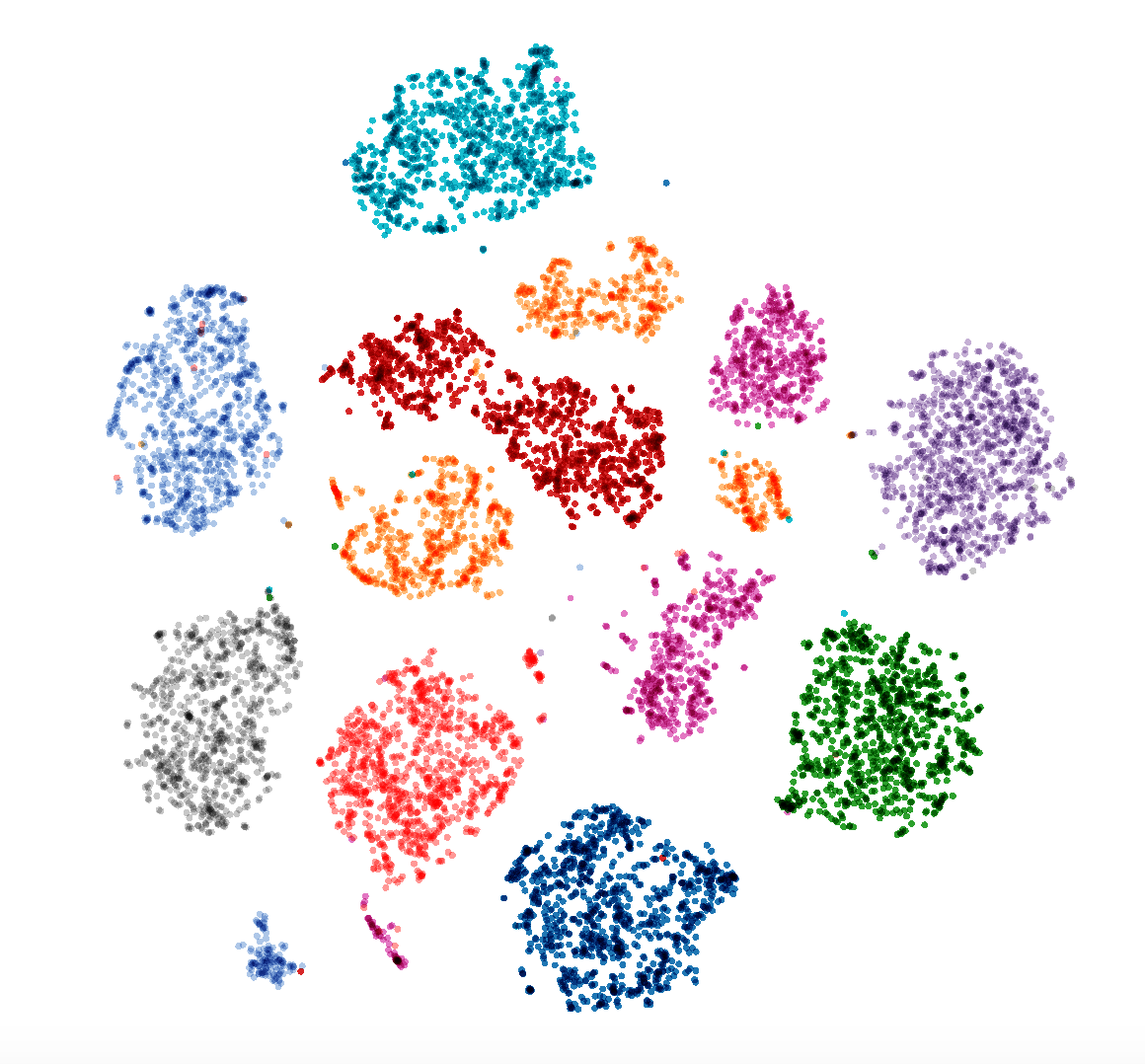}\vspace{-.2cm}
		\caption{MNIST, $\Pi$ model}
		\label{fig:sub3}
	\end{subfigure}
	\begin{subfigure}{.234\textwidth}
		\centering
		\includegraphics[width=\linewidth]{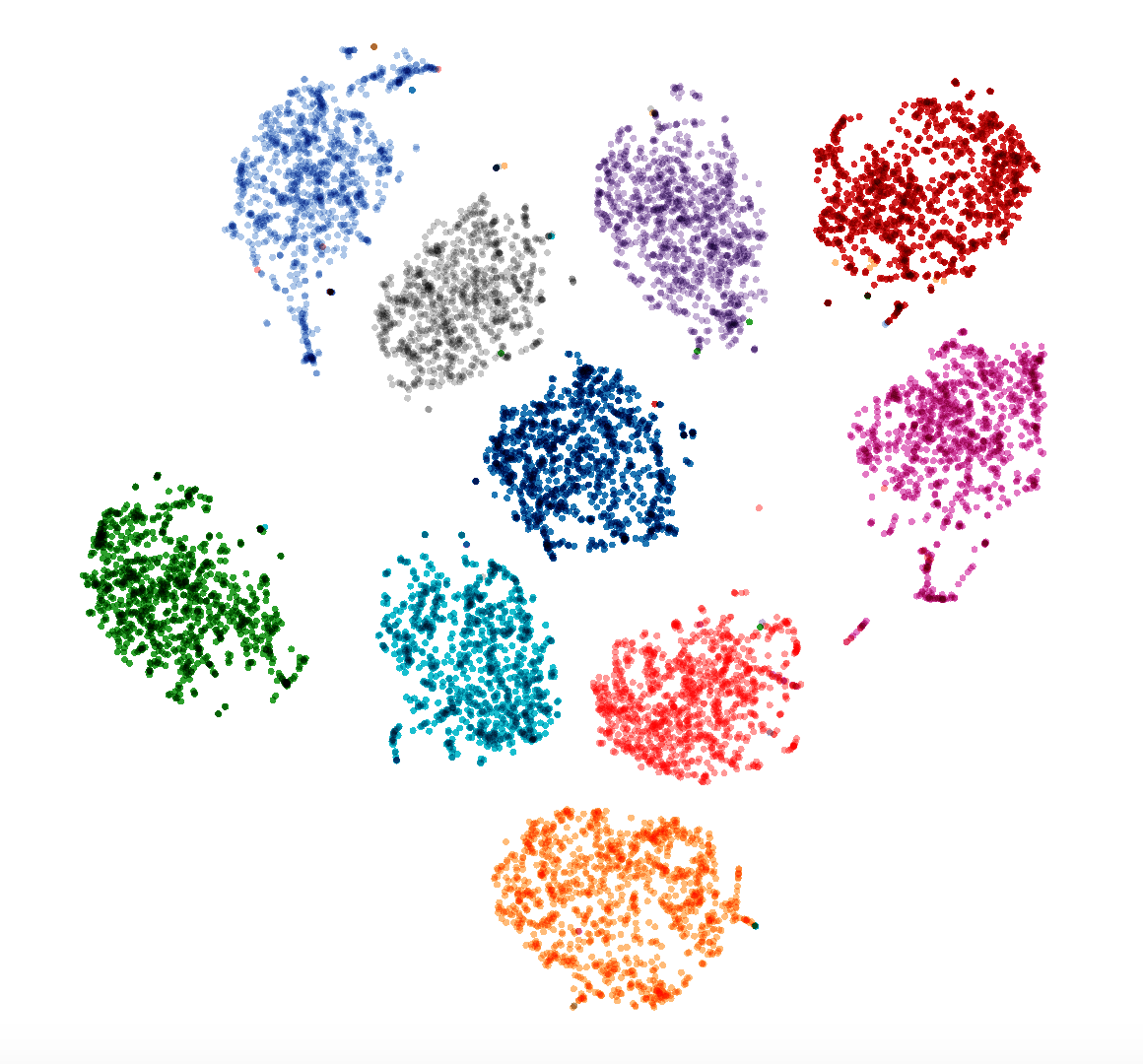}\vspace{-.2cm}
		\caption{MNIST, SNTG}
		\label{fig:sub4}
	\end{subfigure}\vspace{-.2cm}
	\caption{(a, b) are the embeddings of CIFAR-10 test data projected to 2-D using PCA. (c, d) are the 2-D embeddings of MNIST test data using t-SNE. Each color denotes a class. In (b, d) with SNTG, the embeddings of each class form a tight and concentrated cluster. In (c) without SNTG, the cluster of the same class are divided into several parts.}
	\label{fig:embed}\vspace{-.4cm}
\end{figure}
We visualize the embeddings of our algorithm and $\Pi$ model on test data under the same settings (CIFAR-10 with 4000 labels and MNIST with 100 labels, both without augmentation). We implemented it using TensorBoard in TensorFlow~\cite{abadi2016tensorflow}. Fig.~\ref{fig:embed} shows the representations $h(x)\in \mathbb{R}^{128}$ projected to 2 dimension using PCA or tSNE~\cite{maaten2008visualizing}.
The learned representations of our model are more concentrated within clusters and are potentially easier to separate for different classes. The visualization is also consistent with our assumption and analysis.

\subsection{Robustness to noisy labels}
\label{sec:noisy}

We finally show that SNTG can not only benefit from unlabeled data, but also learn from noisy supervision. Following~\cite{laine2016temporal}, we did extra experiments on supervised SVHN to show the tolerance to incorrect labels. Certain percentages of true labels on the training set are replaced by random labels. Fig.~\ref{fig:corrupt_all} shows that TempEns+SNTG retains over 93\% accuracy even when 90\% of the labels are noisy while TempEns alone only obtains 73\% accuracy~\cite{laine2016temporal}. With standard supervised training, the model suffers a lot and overfits to the incorrect information in labels. Thus, our SNTG regularization improves the robustness and generalization performance of the model. Previous work~\cite{reed2014training} also shows that self-generated targets yield robustness to label noise.

\subsection{Feature matching GAN benefits from SNTG}

\label{app:gan}
\begin{figure}[h]\vspace{-.2cm}
	\centering
	\begin{subfigure}{.23\textwidth}
		\centering
		\includegraphics[width=\linewidth]{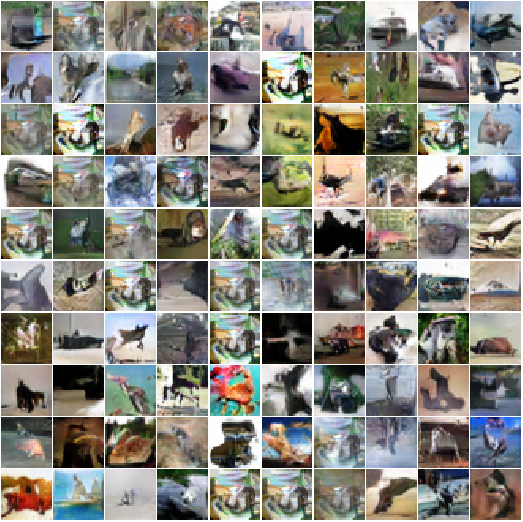}
		\caption{FM GAN}
		\label{fig:fm}
	\end{subfigure}%
	\hfill
	\begin{subfigure}{.23\textwidth}
		\centering
		\includegraphics[width=\linewidth]{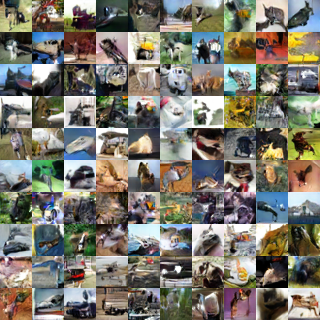}
		\caption{FM GAN+SNTG}
		\label{fig:fmsntg}
	\end{subfigure}\vspace{-.2cm}
	\caption{Comparison of generated images in SSL on CIFAR-10 with FM GAN~\cite{salimans2016improved}, original in their paper ({\em left}) and with our SNTG loss ({\em right}). FM GAN has strange and repeated patterns in the samples. Adding SNTG, the quality and diversity of generated samples are improved.}
	\label{fig:cifar10}\vspace{-.1cm}
\end{figure}
Recently, the feature matching (FM) GAN in Improved GAN~\cite{salimans2016improved} has performed well for SSL but usually generates images with strange patterns. Some works have been done to analyze the reasons~\cite{dai2017good,li2017triple,kumar2017improved}. An interesting finding is that our method can also alleviate the problem.
%It indicates that the FM objective of the generator benefits further from the smooth feature space learned by SNTG.
%In FM GAN, the objective for the generator is defined as\\[-.4cm]
%\begin{eqnarray}
%	\|\mathbb{E}_{x\sim p_{data}}h(x) - \mathbb{E}_{x\sim p_G}h(x)\|_2^2,
%\end{eqnarray}\\[-.5cm]
%which is similar to the neighboring case when
%$W_{ij}=1$ in Eq.~\eqref{marginloss}.
%Although FM GAN performs well in semi-supervised classification, the generated images is not realistic. Some works have been done to analyze the reasons~\cite{dai2017good,li2017triple,kumar2017improved}. We find that our method SNTG can also alleviate the problem.
Fig.~\ref{fig:cifar10} presents the comparison between the samples generated in FM GAN~\cite{salimans2016improved} and FM GAN+SNTG. Apart from improving the generated sample quality of FM GAN, SNTG also reduces the error rate. FM GAN achieves $18.63\%$ on CIFAR-10 with 4000 labels. We regularize the features of unlabeled data using SNTG and observe an improvement to $14.93\%$, which is comparable to the state-of-the-art $14.41\%$ in deep generative models~\cite{dai2017good}.

In FM GAN, the objective for the generator is defined as\\[-.6cm]
\begin{eqnarray}
\|\mathbb{E}_{x\sim p_{data}}h(x) - \mathbb{E}_{x\sim p_G}h(x)\|_2^2,
\end{eqnarray}\\[-.6cm]
which is similar to the neighboring case when
$W_{ij}=1$ in Eq.~\eqref{marginloss}.
In our opinion, SNTG helps shape the feature space better so that the generator could capture the data distribution by matching only the mean of features.
%It is pointed out in~\cite{kumar2017improved} that FM GAN generates moderate fake examples (\ie, the strange patterns as shown in Fig.~\ref{fig:cifar10}). SNTG shapes the feature space that
%	\caption{Comparison of generated images in SSL on CIFAR-10 with feature matching GAN~\cite{salimans2016improved}, original in their paper ({\em left}) and with our SNTG loss ({\em right}). There are strange and repeated patterns in feature matching GAN. Adding SNTG, the quality of generated samples are improved.}

\section{Conclusions and future work}
We present a simple but effective SNTG, which
%, in addition to regularizing the prediction of a single point under perturbation in SSL, now
regularizes the neighboring points on a learned teacher graph. Empirically, it outperforms all baselines and achieves new state-of-the-art results on several datasets. As a byproduct, we also learn an invariant mapping on a low-dimensional manifold. SNTG offers additional benefits such as handling extreme cases with fewer labels and noisy labels.
In future work, it is promising to do more theoretical analysis of our method and to explore its combination with generative models as well as applications to large-scale datasets, \eg, ImageNet with more classes.

\vspace{-1ex}
\section*{Acknowledgements}
\vspace{-1ex}
\small{
The work is supported by the National NSF of China (Nos. 61620106010, 61621136008, 61332007), Beijing Natural Science Foundation (No. L172037), Tsinghua Tiangong Institute for Intelligent Computing, the NVIDIA NVAIL Program and a research fund from Siemens.
}

\clearpage

{\small
	\bibliographystyle{ieee}
	\bibliography{egbib}
}

\clearpage
\appendix	
\section{Experimental setup}
\label{app:1}
\begin{table}
	\centering
	\caption{\label{table:network}
		The network architecture used in all experiments.}
	\begin{tabular}{l}
		\toprule
		Input: $32\times32\times3$ image ($28\times28\times1$ for MNIST)\\
		Gaussian noise $\sigma=0.15$ \\
		$3\times3$ conv. $128$ lReLU ($\alpha=0.1$) same padding\\
		$3\times3$ conv. $128$ lReLU ($\alpha=0.1$) same padding\\
		$3\times3$ conv. $128$ lReLU ($\alpha=0.1$) same padding\\
		$2\times2$ max-pool, dropout 0.5 \\
		$3\times3$ conv. $256$ lReLU ($\alpha=0.1$) same padding\\
		$3\times3$ conv. $256$ lReLU ($\alpha=0.1$) same padding\\
		$3\times3$ conv. $256$ lReLU ($\alpha=0.1$) same padding\\
		$2\times2$ max-pool, dropout 0.5 \\
		$3\times3$ conv. $512$ lReLU ($\alpha=0.1$) valid padding\\
		$1\times1$ conv. $256$ lReLU ($\alpha=0.1$) \\
		$1\times1$ conv. $128$ lReLU ($\alpha=0.1$) \\
		Global average pool $6\times6$ ($5\times5$ for MNIST) $\to 1\times$1\\
		Fully connected $128 \to 10$ softmax\\
		\bottomrule
	\end{tabular}
\end{table}

\begin{table*}[t]
	\caption{Comparision of error rates on MNIST with 600 labels between various classifcal SSL methods.}
	\label{classical}
	\centering
	\begin{tabular}{lllllllllll}
		\toprule
		Methods 	&LGC	&TSVM&LapRLS&LP&LP+$k$NN&DLP& \emph{Embed}NN & MTC &PEA&SNTG (\textbf{ours})\\
		\midrule
		Error (\%) 	&3.96	&4.87		&2.92		&8.57 & 4.27& 2.01&3.42&5.13&2.44&\textbf{0.45}\\
		\bottomrule
	\end{tabular}
\end{table*}
\textbf{MNIST.}
It contains 60,000 gray-scale training images and 10,000 test images from handwritten digits $0$ to $9$. The input images are normalized to zero mean and unit variance.

\textbf{SVHN.}
Each example in SVHN is a $32\times32$ color house-number images and we only use the official 73,257 training images and 26,032 test images following previous work.
The augmentation of SVHN is limited to random translation between $[-2,2]$ pixels.

\textbf{CIFAR-10.}
The CIFAR-10 dataset consists of $32\times32$ natural RGB images from 10 classes such as airplanes, cats, cars and horses. We have 50,000 training examples and 10,000 test examples. The input images are normalized using ZCA following previous work~\cite{laine2016temporal}.
We use the standard way of augmenting the CIFAR-10 dataset including horizontal flips and random translations.

\textbf{CIFAR-100.}
The CIFAR-100 dataset consists of $32\times32$ natural RGB images from 100 classes. We have 50,000 training examples and 10,000 test examples. The preprocession of inputs images are the same to CIFAR-10.

\textbf{Implementation.}
We implemented our code mainly in Python with Theano~\cite{2016arXiv160502688short} and Lasagne~\cite{lasagne}. For comparison with VAT~\cite{miyato2017virtual} and Mean Teacher~\cite{tarvainen2017mean} experiments, we use TensorFlow~\cite{abadi2016tensorflow} to match their settings.
%Code reproducing our results will be available on Github soon after the review.
%at \url{https://github.com/xinmei9322/SNTG} soon.
The code for reproducing the results is available at \url{https://github.com/xinmei9322/SNTG}.

\textbf{Training details.}
In $\Pi$ model and TempEns based experiments, the network architectures (shown in Table~\ref{table:network}) and the hyper-parameters are the same as our baselines~\cite{laine2016temporal}.
We apply mean-only batch normalization with momentum $0.999$~\cite{salimans2016weight} to all layers and use leaky ReLU~\cite{maas2013rectifier} with $\alpha=0.1$.
The network is trained for $300$ epochs using Adam Optimizer~\cite{kingma2014adam} with mini-batches of size $n=100$ and maximum learning rate $0.003$ (exceptions are that TempEns for SVHN uses $0.001$ and MNIST uses $0.0001$). We use the default Adam momentum parameters $\beta_1=0.9$ and $\beta_2=0.999$. Following~\cite{laine2016temporal}, we also ramp up the learning rate and the regularization term during the first 80 epochs with weight $w(t) = \exp\left[-5(1-\frac{t}{80})^2\right]$ and ramp down the learning rate during the last 50 epochs. The ramp-down function is $\exp\left[-12.5(1 - \frac{300- t}{50})^2\right]$.
The regularization coefficient of consistency loss $R_C$ is $\lambda_1=100$ for $\Pi$ model and $\lambda_1=30$ for TempEns (exception is that SVHN with $L=250$ uses $\lambda_1=50$).

For comparison with Mean Teacher and VAT, we keep the same architecture and hyper-parameters settings with the corresponding baselines~\cite{tarvainen2017mean,miyato2017virtual}. Their network architectures are the same as shown in Table~\ref{table:network} but differ in several hyper-parameters such as weight normalization, training epochs and mini-batch sizes, which are detailed in their papers. We just add the SNTG loss along with their regularization $R_C$ and keep other settings unchanged as in their public code.

In all our experiments, the margin $m$ in $R_S$ %Eq.~\eqref{marginloss}
is set to $m = 1$ if we treat $\|h(x_i) - h(x_j) \|^2$ as a distance averaged by the feature dimension $p$.
%when $\|h(x_i) - h(x_j) \|^2$ is averaged by the feature dimension $p$.
We sample half the number of mini-batch size pairs of $(x_i, x_j)$ for computing $\ell_G$, \eg, $s = 50$ for mini-batch size $n = 100$. The regularization coefficient $\lambda_2$ of SNTG loss $R_S$ is set to $\lambda_2 = k\lambda_1$ where $k$ is the ratio of $\lambda_2$ to $\lambda_1$ (\ie, the regularization coefficient of consistency loss $R_C$). $k$ is chosen from $\{0.2, 0.4, 0.6, 1.0\}$ using the validation set and we use $k=0.4$ for most experiments by default.

\textbf{Training time.} SNTG does not increase the number of neural network parameters and the runtime is almost the same to the baselines, with only extra 1-2 seconds per epoch (the baselines usually need 100-200 seconds per epoch on one GPU).

\textbf{Synthetic benchmarks.}
The synthetic dataset experiments adopt the default settings for $\Pi$ model~\cite{laine2016temporal} except for $0.001$ maximum learning rate and $500$ training epochs. We use weight normalization~\cite{salimans2016weight} and add Gaussian noise to each layer.

\section{Rethinking $\Pi$ model objective}
\label{app:rethink}
In $\Pi$ model~\cite{laine2016temporal}, the consistency loss is defined in Eq.~({\color{red}2}) %\eqref{equ:3}
where the teacher model shares the same parameter with the student model $\theta' = \theta$. Suppose $f(x)\in [0,1]^K$, the consistency loss of $\Pi$ model is
\begin{eqnarray*}
	R_C(\theta, \mathcal{L}, \mathcal{U}) = \sum_{i=1}^{N}\mathbb{E}_{\xi',\xi}\|  f(x_i;\theta, \xi') - f(x_i;\theta, \xi)\|^2 ,
\end{eqnarray*}
$\xi'$ and $\xi$ are i.i.d random noise variables, $\xi',\xi \sim p(\xi)$, then we have $\mathbb{E}_\xi f(x_i;\theta,\xi) = \mathbb{E}_{\xi'} f(x_i;\theta,\xi')$ and $\mathbb{E}_\xi \|f(x_i;\theta,\xi)\|^2 = \mathbb{E}_{\xi'} \|f(x_i;\theta,\xi')\|^2$
\begin{eqnarray*}
	R_C \!\!&\!=\!&\!\! 2\sum_{i=1}^{N}\mathbb{E}_{\xi}\|f(x_i; \theta, \xi)\|^2 -\| \mathbb{E}_{\xi}f(x_i;\theta, \xi)\|^2\\
	\!\!&\!=\!&\!\! 2\sum_{i=1}^{N}\sum_{k=1}^{K}\mathrm{Var}_{\xi}\left[f(x_i;\theta, \xi)\right]_k
\end{eqnarray*}
where $[\cdot]_k$ is $k$-th component of the vector.

Then minimizing $R_C$ is equivalent to minimizing the sum of variance of the prediction each dimension. Similar idea of variance penalty was exploited in Pseudo-Ensemble~\cite{bachman2014learning}. If a data point is near the decision boundary, it is likely to has a large variance since its prediction might alternate to another class when some noise is added. Minimizing the variance explicitly penalizes such alternation behavior of training data.

\section{Comparison to classical SSL methods}
As mentioned in Section~{\color{red}2}, our method is different from classical graph-based SSL methods in many important aspects such as the construction of the graph and how to use it.

Table~\ref{classical} is a comparion with several classical methods: (1) Label propagation (LP)~\cite{zhu2002learning}; (2) A variant of LP on $k$NN structure(LP+$k$NN)~\cite{subramanya2010efficient}; (3) Local and Global Consistency (LGC)~\cite{zhou2004learning}; (5) Transductive SVM (TSVM)~\cite{joachims1999transductive}; (6) LapRLS~\cite{belkin2006manifold}; (7) Dynamic Label propagation (DLP)~\cite{wang2013dynamic}. The results of (1)-(7) are cited from~\cite{wang2013dynamic}. We also compare with the best reported results in previously mentioned works: (8) \emph{Embed}NN~\cite{weston2008deep}; (9) the Manifold Tangent Classifier (MTC)~\cite{rifai2011manifold}; (10) Pseudo-Ensemble~\cite{bachman2014learning}.

While the classical graph-based methods (\eg, LP, DLP and LapRLS) were the leading paradigms, with the resurgence of deep learning, recent impressive results are mostly from deep learning based SSL methods, while classical methods fall behind on performance and scalability. Furthermore, they have no reported results on challenging natural image datasets, \eg, SVHN, CIFAR-10. Only one overlap is MNIST, see Table~\ref{classical} for comparison. We show that our method SNTG surpasses these classical methods by a large margin.

\section{Significance test  of the improvements.}

 Table~\ref{ttest} shows the independent two sample T-test on the error rates of baselines and our method. All the P-values are less than significance level $\alpha=0.01$. It indicates that the improvements of SNTG are significant.
\begin{table}[h]
	\vspace{-.2cm}
	\caption{T-test. The top rows are the experiments without augmentation and the bottom rows are with augmentation.}
	\vspace{-.3cm}
	\centering
	\resizebox{.49\textwidth}{!}{ 	
		\begin{tabular}{p{0.83\linewidth}p{0.165\linewidth}p{0.215\linewidth}}
			\toprule
			Datasets \& Methods				&T-statistic				 	&P-value\\
			\midrule
			{\small MNIST (L=20)\qquad\, $\Pi$ model v.s. $\Pi$+SNTG} & 20.00227 &9.07043e-09\\
			{\small MNIST (L=100)\qquad $\Pi$ model v.s. $\Pi$+SNTG} & 4.34867 &0.000387026\\
			{\small SVHN (L=1000) \quad\, VAT+Ent v.s. VAT+Ent+SNTG} & 4.08627 		&0.002732236\\
			{\small CIFAR-10 (L=4000) VAT+Ent v.s. VAT+Ent+SNTG}& 5.90681 	&0.000227148\\
			\midrule
			{\small SVHN (L=250)\qquad\, $\Pi$ model v.s. $\Pi$+SNTG} & 12.31365 & 3.32742e-10\\
			{\small SVHN (L=500)\quad\,\,\,\,\,\, TempEns v.s. TempEns+SNTG} & 7.52909 & 3.58188e-05\\
			{\small CIFAR-10 (L=1000) TempEns v.s. TempEns+SNTG}& 12.81875 &1.73155e-10\\
			{\small CIFAR-10 (L=2000) TempEns v.s. TempEns+SNTG}& 11.80608 &6.55694e-10\\
			{\small CIFAR-10 (L=4000) VAT+Ent v.s. VAT+Ent+SNTG}& 5.81409 &0.000254937\\
			\bottomrule
		\end{tabular}
	}
	\label{ttest}
	\vspace{-.5cm}
\end{table}

\end{document}